\pdfoutput=1
\documentclass[lettersize,journal]{IEEEtran}
\usepackage{amsmath,amsfonts}
\usepackage{algorithmic}
\usepackage{algorithm}
\usepackage{array}
\usepackage[caption=false,font=normalsize,labelfont=sf,textfont=sf]{subfig}
\usepackage{textcomp}
\usepackage{stfloats}
\usepackage{url}
\usepackage{verbatim}
\usepackage{graphicx}
\usepackage{cite}
\begin{document}

\title{Transfer Learning and Vision Transformer based State-of-Health prediction of Lithium-Ion Batteries}

\author{Pengyu Fu, Liang Chu, Zhuoran Hou, Jincheng Hu, ~\IEEEmembership{Student Member,~IEEE}, Yanjun Huang, ~\IEEEmembership{Member,~IEEE}, and Yuanjian Zhang, ~\IEEEmembership{Member,~IEEE}
\thanks{(Corresponding author: Yuanjian Zhang.)}
\thanks{Pengyu Fu, Liang Chu and Zhuoran Hou are with the Collge of Automotive Engineering, Jilin University, Changchun 130022, China (e-mail: fupy20@mails.jlu.edu.cn; chuliang@jlu.edu.cn; houzr20@mails.jlu.edu.cn).}

\thanks{Jincheng Hu and Yuanjian Zhang are with the     Department of Aeronautical and Automotive Engineering, Loughborough University, Loughborough, U.K (e-mail: jincheng.hu2020@outlook.com; y.y.zhang@lboro.ac.uk).}

\thanks{Yanjun Huang is with the School of Automotive Studies, Tongji University, Shanghai, China. (e-mail:huangyanjun404@gmail.com).}}

\markboth{Journal of \LaTeX\ Class Files,~Vol.~14, No.~8, August~2021}%
{Shell \MakeLowercase{\textit{et al.}}: A Sample Article Using IEEEtran.cls for IEEE Journals}

\IEEEpubid{0000--0000/00\$00.00~\copyright~2021 IEEE}

\maketitle

\begin{abstract}
In recent years, significant progress has been made in transportation electrification. And lithium-ion batteries (LIB), as the main energy storage devices, have received widespread attention. Accurately predicting the state of health (SOH) can not only ease the anxiety of users about the battery life but also provide important information for the management of the battery. This paper presents a prediction method for SOH based on Vision Transformer (ViT) model. First, discrete charging data of a predefined voltage range is used as an input data matrix. Then, the cycle features of the battery are captured by the ViT which can obtain the global features, and the SOH is obtained by combining the cycle features with the full connection (FC) layer. At the same time, transfer learning (TL) is introduced, and the prediction model based on source task battery training is further fine-tuned according to the early cycle data of the target task battery to provide an accurate prediction. Experiments show that our method can obtain better feature expression compared with existing deep learning methods so that better prediction effect and transfer effect can be achieved.
\end{abstract}


\section{Introduction}
\IEEEPARstart{W}{ith} the continuous increase in car ownership, the demand for non-renewable energy is increasing, and the environmental pollution caused by exhaust emissions is also becoming increasingly serious \cite{XIE2021119975, AGARWAL2021110624, KILKIS2018164}. To solve these problems, the electric vehicle (EV) with power batteries to completely or partially replace fuel as the power source has become the research hotspot in recent years \cite{JAGUEMONT2020101551}. Lithium-ion battery (LIB) is the most widely used battery type in EV because of its high energy density, high output power and long cycle life \cite{DENG201991,8536873}. With excellent performance, LIBs that work outside of safe operation area (SOA) for a long time may affect the performance of the battery or even cause serious safety accidents \cite{LIU2022103910}. The battery management system (BMS) can monitor and regulate the charge and discharge of the battery to ensure performance and safety, and State of health (SOH) prediction is a very important function in BMS \cite{WANG2020110015}. Accurate prediction of SOH can provide a reference for battery control strategy and effectively avoid problems caused by battery working outside SOA \cite{2022zhang}. However, the working environment of LIBs is very variable, and the ageing mechanism is complex \cite{LI2020117852,9000971, SONG2020101836}. SOH prediction of LIBs is a challenging problem.

 In recent years, researchers have proposed many SOH prediction methods, which can be divided into direct measurement methods, model-driven methods, and data-driven methods.

According to the treatment, subtypes of the direct measurement method, including destructive and non-destructive measurement methods, are widely adopted. The destructive measurement method needs to disassemble the battery and test the positive electrode, negative electrode, separator, electrolyte or gas to obtain the ageing condition of the battery \cite{PETZL201480}. Such as X-ray diffraction (XRD) \cite{zhang2016255}, X-ray photoelectron spectroscopy (XPS) \cite{SCIPIONI2018454}, and gas chromatography \cite{2020gas}. The destructive measurement method can intuitively check the internal state of the batteries, with high prediction accuracy and reliable diagnosis results.  It is also convenient to analyze the ageing mechanism of LIBs. However, the measurement process destroyed the structure of the batteries, making the battery unusable. Nevertheless, the non-destructive measurement method prevents batteries from damaging the battery framework. Such as Acoustic Emission Detection(AE) \cite{s21030712}, Hybrid Pulse Power Characteristic (HPPC) \cite{7013631} and Electrochemical Impedance Spectroscopy (EIS) \cite{MEDDINGS2020228742}. To be specific, the SOH prediction accuracy of non-destructive measurement methods is directly affected by the sampling accuracy of data. Therefore, the measurement environment of the non-destructive measurement method is demanding, and it is difficult to be applied to the actual scene.
\IEEEpubidadjcol

Model-driven methods can be divided into Electrical Model (EM), Equivalent Circuit Model (ECM) and Reduced-order Simplified Model(RSM). EM describes the charging and discharging behaviour of the battery based on the electrochemical principle, and studies the influence of different ageing factors on the state variables in the ageing process \cite{2020simon,8001575}. EM has strong applicability, strong explanatory power and high prediction accuracy, but these advantages are based on the accuracy of modelling. The high-precision model requires complex partial differential equations to build the model, and it is difficult to obtain the fine internal parameters of the battery. This makes it difficult to establish a high-precision model, which affects the effect of EM. ECM uses different combinations of simple electrical devices to simulate the charge and discharge behaviour of LIBs \cite{9254859, TRAN2021103252}. Therefore, ECM has a simple structure, high calculation efficiency and few identification parameters.
However, ECM cannot capture the internal electrochemical state of the battery, which restrains the upper bound of the performance of advanced BMS. At the same time, the accuracy of ECM is not only affected by the model structure but also limited by the model parameter identification algorithm \cite{LAI20191057}. With the development of order reduction technology \cite{jokar201644}, RSM simplifies the strong linear relationship in EM by effective mathematical means based on EM and balances the interpretability and computational efficiency to obtain similar prediction accuracy with fewer parameters \cite{2020cen, LAI2020135239}. However, RSM needs to maintain acceptable model accuracy while reducing computational complexity. Oversimplification leads to RSM only applicable to some working conditions \cite{MEHTA2021138623}. In addition, the rationality of RSM needs a large number of experimental data to verify. These are the problems that RSM needs to solve urgently.

Data-driven methods do not depend on the mechanism of the battery and are flexible in use. Data-driven methods can learn from data, and make predictions with minimal human intervention. But data-driven methods need a large amount of training data. With the emergence of various battery datasets and the development of computer technology, more and more researchers focus on data-driven methods of SOH prediction. The data-driven methods for SOH prediction can be divided into the Difference Analysis (DA) method, basic machine learning method and deep learning method. DA is developed based on a simple mapping relationship between the differential features of the battery and its attenuation capacity. The features of the battery in the charge and discharge cycle contain information on battery ageing, but the change in the features is generally small. After feature differentiation, it is easier to observe and process. Zhang et al. \cite{ZHANG2020228740} proposed a model-free SOH prediction method integrating Coulomb Counting and Differential Voltage Analysis (DVA). The voltage axis was replaced by the SOC axis, and two SOC feature points were identified. Finally, the measurement time and average SOC corresponding to these two feature points were substituted into the derivation formula to estimate SOH. Erik et al. \cite{9328130} used the Incremental Capacity Analysis (ICA) method to obtain the characteristic peaks and valleys of the IC curve to predict the SOH of the battery. Wang et al. \cite{9212409} used Differential Thermal Analysis (DTV) to process battery data and extract health factors from peak positions, peaks and valleys of DTV curves. Although the DA method has high calculation efficiency and prediction accuracy in SOH prediction, the selection of differential intervals and characteristic points has a great impact on the results of the DA method. It is required that the test data have high quality, so it is difficult to apply the DA method online.

Different from DA methods, basic machine learning methods focus on data features. Basic machine learning methods usually use the features extracted from experimental data to estimate SOH, so complete experimental data is not required. Common basic machine learning methods include Support Vector Regression (SVR) \cite{ZHANG2022121986,LI2022104215}, Gaussian Process Regression (GPR) \cite{9647910,2021zhou}, Relevance Vector Machine (RVM) \cite{electronics10121497,app9091890}. The main process of basic machine learning: 1. Obtain various battery data through sensors. 2. After preprocessing, feature extraction and feature selection, the feature expression of the data is obtained. 3. Complete the inference and prediction of results through features. Among them, proper feature expression plays a key role in the accuracy of the final algorithm. Therefore, extracting good feature expression is the focus of basic machine learning methods. Guo et al. \cite{GUO2021102372} used the IC curve of the constant current process to obtain the characteristics of four health indicators and used SVR to connect the health indicators with the SOH of the battery. Wang et al. [2021wang] introduced the minimum redundancy maximum correlation (mRMR) algorithm to select the best feature set and combined it with multi-kernel RVM to obtain higher prediction accuracy. It can be seen that feature extraction takes a lot of calculation and testing work in basic machine learning. At the same time, feature extraction is mostly defined manually, so it is difficult to identify the features suitable for different experimental conditions.

The concept of deep learning comes from the research of artificial neural networks (ANN). Deep learning combines low-level features with multiple hidden layers to form more abstract high-level features to discover the distributed feature representation of data. Therefore, the deep learning method optimizes and shortens the process of data analysis. Shen et al. \cite{SHEN2019100817} first tried to introduce deep learning into the SOH prediction task, input the collected battery data into Convolutional Neural Network (CNN) in the form of the matrix, and obtained good prediction results. In addition, the Long Short Term Memory network (LSTM) is a deep learning network specialized in learning long-term dependence. LSTM has an internal structure that can mine the ageing feature of the battery from the historical charge and discharge cycle data. In the case of noise interference, LSTM can provide robust and flexible results. However, the battery features extraction ability of LSTM is insufficient, and it is necessary to select certain ageing features from each charge and discharge cycle as input. Tan et al. \cite{8870196} extracted nine ageing features on the voltage curve of the CC step as input of LSTM, and used Gray Relational Analysis (GRA) to verify the validity of ageing features. 

It can be seen that the effect of the deep learning method still depends on the feature extraction ability of the model, but the existing deep learning methods have insufficient ability to extract battery features. This is because many deep learning algorithms often make some assumptions about learning problems, which are called inductive bias. The inductive bias of LSTM is sequentially and time variance and the inductive bias of CNN is locality and shift-invariance. Strong inductive bias makes it possible to achieve high performance even with fewer data, but the mismatch between inductive bias and target task will affect the acquisition of feature expression and limit the performance of the model. To alleviate the above problems, this paper proposes a SOH prediction model based on Vision Transformer (ViT). The model combines ViT and the Full Connection (FC) layer. The Self-attention layer of ViT is global, with the minimum inductive bias, ensuring the flexibility and performance of feature acquisition. As the regression layer, FC maps the feature expression obtained by the ViT to SOH, greatly reducing the influence of the feature position on the regression. At the same time, the Transfer 
Learning (TL) is introduced to transfer the information of the source task to reduce the model training cost and improve prediction accuracy. The main work and contributions of this paper are as follows:

\begin{enumerate}
\item{The advanced deep learning method ViT is used, which does not need to manually extract features and automates the process of feature learning. When extracting battery cycle features, ViT can learn more global features, thus achieving higher prediction accuracy.}
\item{Through the multi-layer transformer encoder structure, ViT can transform shallow battery features into higher and more abstract feature representations, and mine highly representative features in battery data. When only current, voltage and temperature are used, Vit can also obtain good SOH prediction results.}
\item{Transfer learning is introduced. The hidden layer parameters of ViT are frozen to capture similar battery data feature expressions. At the same time, fine-tune the hidden layer parameters of FC to fit the SOH under different working conditions. It improves the prediction accuracy of the battery under unknown conditions and significantly reduces the experimental cost and training costs.}
\end{enumerate}

The rest of this paper is divided into five parts. Section \uppercase\expandafter{\romannumeral2} introduces the data used in this paper. Section \uppercase\expandafter{\romannumeral3} introduces the principle of the method and the specific structure of the model. Section \uppercase\expandafter{\romannumeral4} presents and explains the validation results. Section \uppercase\expandafter{\romannumeral5} summarizes some conclusions and future research directions.

\section{Experimental Data}
In this section, the definition of SOH, the source of original data, the preprocessing of battery data, and the composition of the dataset are explained in detail.
\subsection{Definition of SOH}
The SOH of the battery represents a certain stage of the battery life. It evaluates the health level of the current specific performance compared to the new state. However, there is no uniform definition of SOH. In the existing studies, most researchers use characterization parameters such as capacity or impedance to evaluate the health status \cite{19,20}. The other researchers use relevant ageing mechanism parameters to monitor the recyclable lithium ions \cite{21}, or the solid-phase diffusion time of lithium ions in the positive electrode \cite{22}, to evaluate the health status of the battery in this way.

To quantify the battery SOH, two traditional definitions based on capacity and impedance are usually used, and the formula is as follows:
\begin{equation}
\label{eq_1}
SO{{H}_{E}}=\frac{{{C}_{aged}}}{{{C}_{fresh}}}\times 100 \%
\end{equation}
\begin{equation}
\label{eq_2}
SO{{H}_{P}}=\frac{{{R}_{EOL}}-{{R}_{aged}}}{{{R}_{EOL}}-{{R}_{fresh}}}\times 100\%
\end{equation}
where ${C_{fresh}}$ refers to the nominal capacity at a specific charging rate when the battery is in the initial state, ${{C}_{aged}}$ refers to the ageing capacity measured at a specific time, ${{R}_{fresh}}$ refers to the initial internal resistance at the initial state, ${{R}_{EOL}}$ refers to the internal resistance at the end of the life, and ${{R}_{aged}}$ refers to the ageing internal resistance measured or estimated at a specific time. Equation \ref{eq_1} is to quantify SOH by the capacity of the battery, and the energy storage capacity of the battery is the main concern. Equation \ref{eq_2} is to quantify SOH by the impedance of the battery, and the power performance is the main concern.

The internal resistance can be measured by different methods (such as EIS \cite{MEDDINGS2020228742} and HPPC \cite{7013631}), but the measurement is highly sensitive to experimental conditions. The capacity is a value directly measured by the Coulomb Counting method under constant current charging / discharging conditions. Although capacity still depends on measurement parameters such as current and temperature, it is considered a more direct descriptor for SOH prediction. The existing literature on SOH prediction is quite extensive but mainly focuses on capacity prediction. Therefore, this paper also uses capacity as the evaluation parameter of SOH.

\subsection{Source of data}
\begin{table}[!t]
\caption{Manufacturing parameters of battery}
\label{table1}
\centering
\begin{tabular}{|c||c|}
\hline
Properties & \\
\hline
Cathode material & Nickel cobalt aluminium (NCA)\\
Anode material & Graphite/silicon\\
Shape & Cylinder\\
Nominal battery capacity  & 4800mAh\\
Nominal battery voltage & 3.6V\\
Battery weight & 0.08g\\
Battery volume & 0.1L\\
Voltage range & 2-4.3V\\
\hline
\end{tabular}
\end{table}
The power batteries used by vehicles can be divided into three types based on applications, including high-energy batteries suitable for EVs, high-power batteries suitable for Hybrid Electrical Vehicles (HEVs), and batteries with high-power and high-energy performance suitable for plug-in hybrid vehicles (PHEVs). 12 commercial 21700 LIBs of the same type are used in this paper. The specific model is not announced due to the confidentiality agreement. The battery is a high-energy battery, and the basic parameters of the battery are shown in Table \ref{table1}.

During the experiment, all batteries were placed in the climate chamber. The battery ageing test is realized by repeated charge and discharge cycles. The constant current constant voltage (CC-CV) mode was used during charging. The 12 batteries are charged and discharged according to the following process:
\begin{enumerate}
\item{The temperature of the climate chamber is set to ${{T}_{c}}$. The batteries need to be stored in the climate chamber under the temperature ${{T}_{c}}$ for more than 3 hours to reach the thermal equilibration.}
\item{The charging rate (C-rate) in CC step is set to ${{I}_{c}}$, and the cut-off voltage of CC step is $V_{max}$. When the charging voltage reaches $V_{max}$, the CC step is ended.}
\item{After the CC step is completed, immediately switch to the CV step to continue charging until the charging current drops to ${{I}_{min}}$. Thus, the charging phase is ended.}
\item{After charging, leave the battery for one hour to reset the temperature. The battery starts discharging at a constant current and the discharge rate is ${{I}_{d}}$ until the voltage drops to $V_{min}$.}
\end{enumerate}

\begin{figure}[!t]
\centering
\includegraphics[width=2.6in]{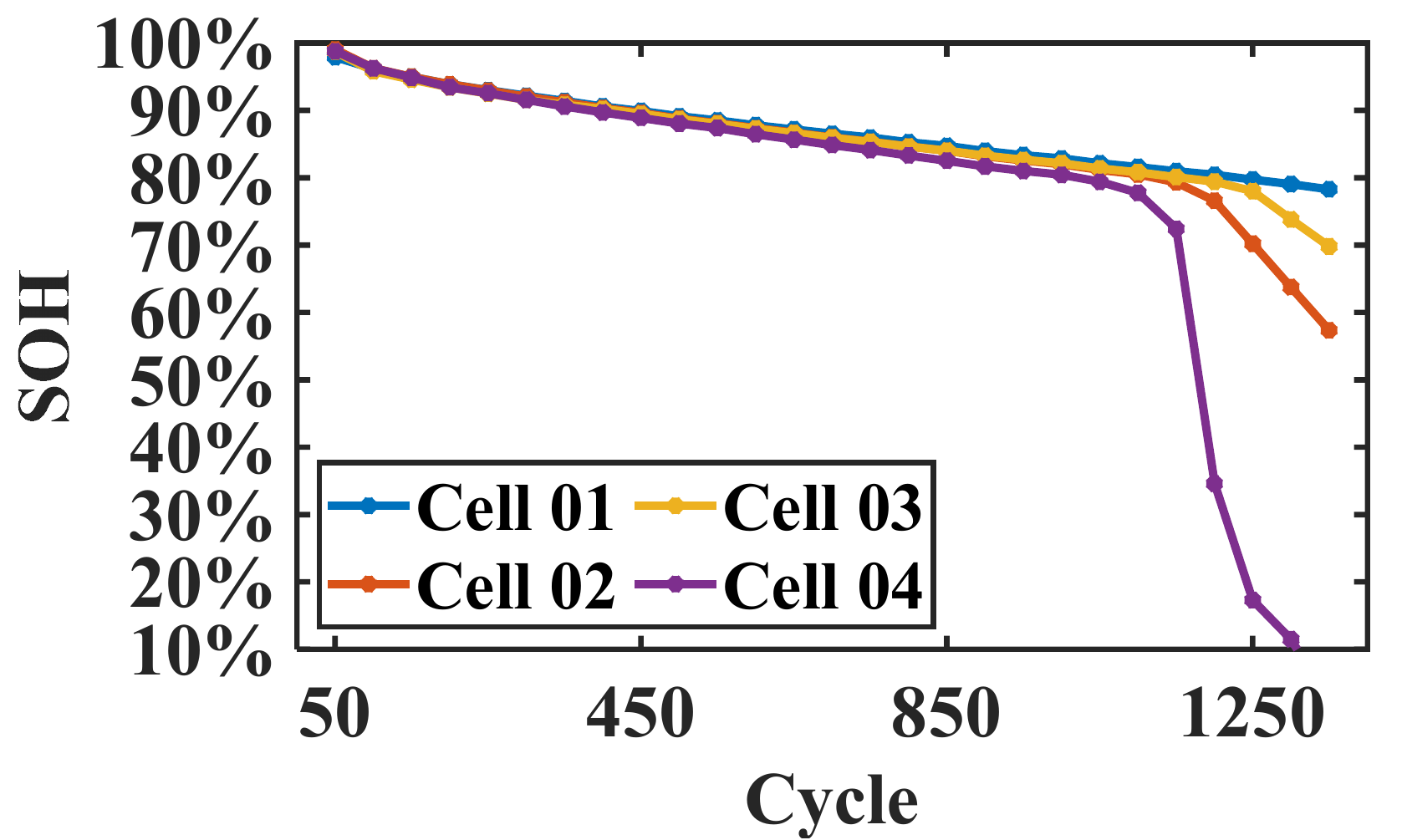}
\caption{Aging curve of the battery dataset.}
\label{fig1}
\end{figure}

In the dataset used in this paper, under the condition that ${{T}_{c}}$, $V_{max}$ and $V_{min}$ are unchanged, 12 different working environments are obtained by changing ${{I}_{c}}$ and ${{I}_{d}}$. The initial capacity test shall be conducted before the ageing cycle test of the battery to determine the actual rated capacity of each battery. Then, the periodic capacity test is conducted at the interval of every 50 full charge and discharge cycles to obtain The initial capacity test shall be conducted before the ageing test of the battery to determine the actual rated capacity of each battery. Then, the periodic capacity test is conducted at the interval of every 50 full charge-discharge cycles (FEC) to obtain the track of battery ageing as shown in Fig \ref{fig1}.

\subsection{Data preprocessing}
In the battery experiment, record the data that can be directly collected, such as current, voltage, temperature, etc. The working environment of a power battery is complex, the discharge process is highly random, and the charging process has a fixed charging law, which is generally consistent. Therefore, the SOH prediction based on the battery data in the charging process can be applied in the actual working conditions. Among all the battery data, the voltage is most easily measured, so the voltage is used as a reference to process other data. The process is as follows:
\begin{enumerate}
\item{According to the collected data points, the time change curve of any battery data can be obtained.}
\item{In CC step, a fixed voltage segment of the lowest voltage ${{V}_{low}}$ and the highest voltage ${{V}_{high}}$ is defined, and the charging time period $\left[ {{t}_{low}},{{t}_{high}} \right]$ corresponding to the voltage segment can be obtained.}
\item{The change curve of any battery data in this time period $\left[ {{t}_{low}},{{t}_{high}} \right]$ can be obtained.}
\item{Finally, the battery data in this time period $\left[ {{t}_{low}},{{t}_{high}} \right]$ is discretized into ${{L}_{V}}$ points. Thus, each type of battery data can be processed into a data vector with a fixed length ${{L}_{V}}$.}
\end{enumerate}

\subsection{Composition of datasets}
After obtaining the data vector of the battery data, the datasets are formed according to the following steps:
\begin{enumerate}
\item{The input data is a matrix spliced by the battery data vector, defined as $X=\left[DV_{1}, \ldots, DV_{n}\right]$, $DV_{n}$ is the $n$-th battery data vector used as input, $X\in{{\mathbb{R}}^{n\times{{L}_{V}}}}$. Each partial charge cycle has a corresponding discharge capacity, which is used to calculate the SOH of the battery as the target output of the model $Y$. Therefore, a set of data $\left[X, Y\right]$ can be obtained for each charge and discharge cycle to train the SOH prediction model.}
\item{Cell 02 and Cell 07 are randomly selected as the target task batteries, and the remaining ten batteries are selected as the source task batteries. The source task batteries can be considered as known working condition batteries that have collected all ageing stages in the battery experiment, while the target task batteries can be considered as new batteries with unknown working conditions. Although the ageing features of the batteries in these two tasks are not completely consistent, there is a similar mapping relationship between the battery input data $X$ and the target output SOH value $Y$.}
\item{All the data of the source task batteries are known, and a fixed proportion of data is randomly selected as the training set of the source task. This proportion is the training set ratio, which is recorded as ${{R}_{t}}$. The remaining data in the source task batteries are the test set of the source task.}
\item{All the data of the target task batteries are unknown, so from the new stage of each battery, the data of the first $c$ cycles are collected as the training set of the target task. And the remaining data of the target task batteries are the test set of the target task.}
\end{enumerate}

\section{Methodologies}
In this section, a SOH prediction model based on ViT \cite{2021An} and FC is proposed. The function of ViT is to divide the battery data $X$ into patches and capture the feature expression of the battery data by calculating the attention weight of the patches. The function of FC is to map the feature expression captured by ViT to the SOH prediction value of the battery. During the transfer learning process, the hidden layer parameters of ViT are frozen to capture similar battery data feature expressions. At the same time, fine-tune the hidden layer parameters of FC to fit SOH under different working conditions.

\subsection{Overall Architecture of ViT-FC}
Transformer \cite{trans} uses the Self-attention to calculate the attention weights of all inputs, and has a strong ability to extract global features. Therefore, it has achieved great success in the field of Natural Language Processing (NLP). And the ability to capture global features is lacking in CNN and LSTM. However, the model complexity of Transformer is the square level of input data. When the input data increases, Floating Point OPerations (FLOPs) will increase, affecting the efficiency of training and fine-tuning. This limits the application of Transformer in SOH prediction.

\begin{figure}[!t]
\centering
\includegraphics[width=3.4in]{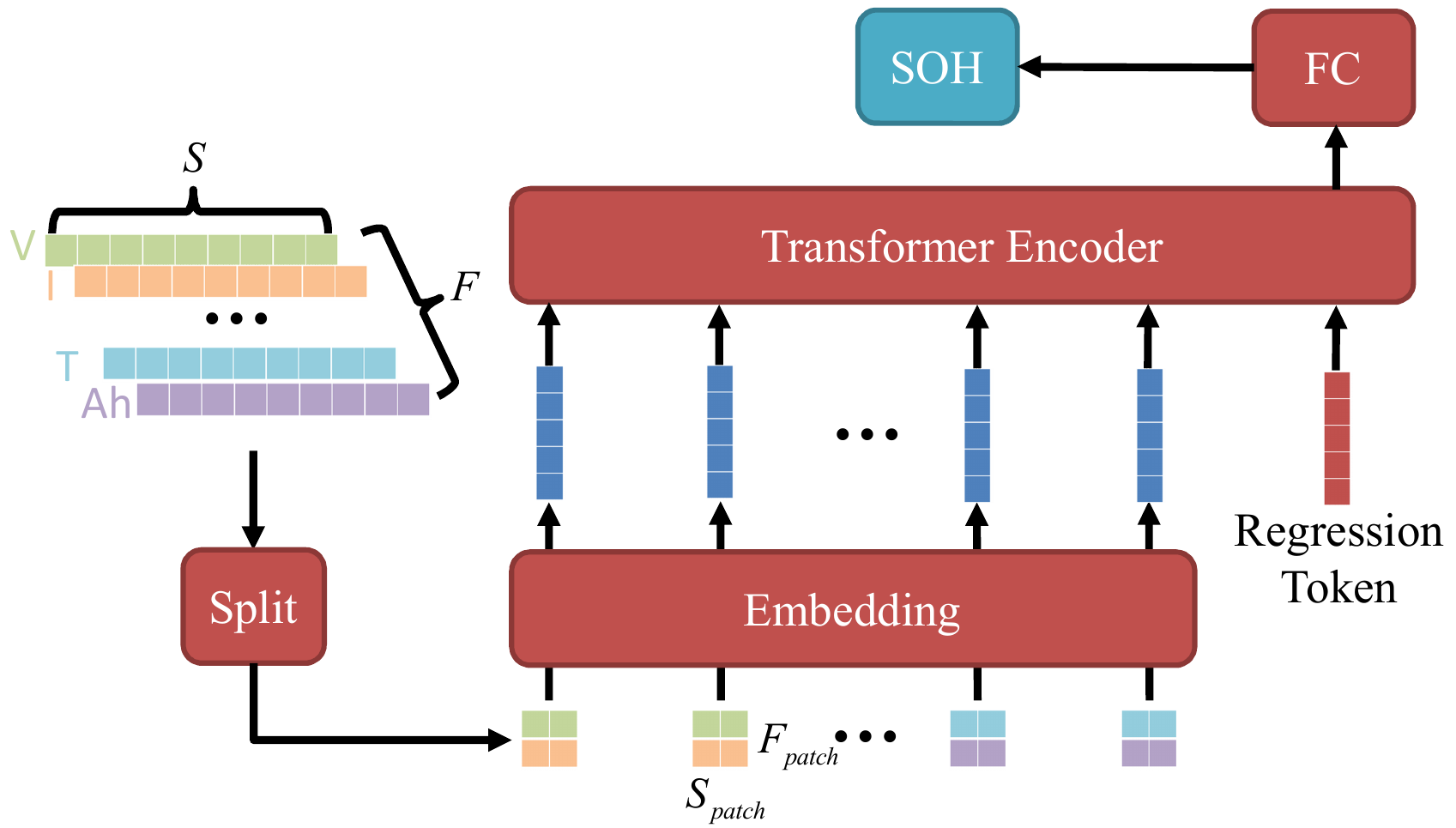}
\caption{Overall Architecture of ViT-FC.}
\label{fig2}
\end{figure}

To apply Transformer to SOH prediction, ViT splits the entire data matrix into several patches. All of the patches are treated the same way as tokens (words). After all, tokens are rearranged into a sequence, the sequence is input into the Transformer Encoder to extract features. The ViT model can be divided into two parts. First, the input battery data matrix is embedded into a sequence with position embedding. Then the cycle features of the battery are captured through Transformer Encoder. In addition, unlike the basic ViT, the final classifier is replaced with an FC layer for regression. And it maps the features extracted by ViT to the predicted SOH. The overall structure of ViT-FC is shown in Fig. \ref{fig2}.

\subsubsection{Embedding of battery data}
Data matrix embedding is mainly divided into two parts, one is patch embedding, the other is position embedding.

As shown in the left half of Fig. \ref{fig2}, the battery data matrix is divided into patches of the same size. $S$ is the discretization granularity of battery data, and $F$ is the number of input battery data. ${{S}_{patch}}$ is the discretized data included in the patch, and ${{F}_{patch}}$ is the type of battery data included in the patch. Therefore, the input battery data is divided into $\left( S/{{S}_{patch}} \right)\times \left( F/{{F}_{patch}} \right)$ patches. The number of patches is the length of the sequence after embedding and is recorded as $L$.

\begin{figure}[!t]
\centering
\includegraphics[width=3.4in]{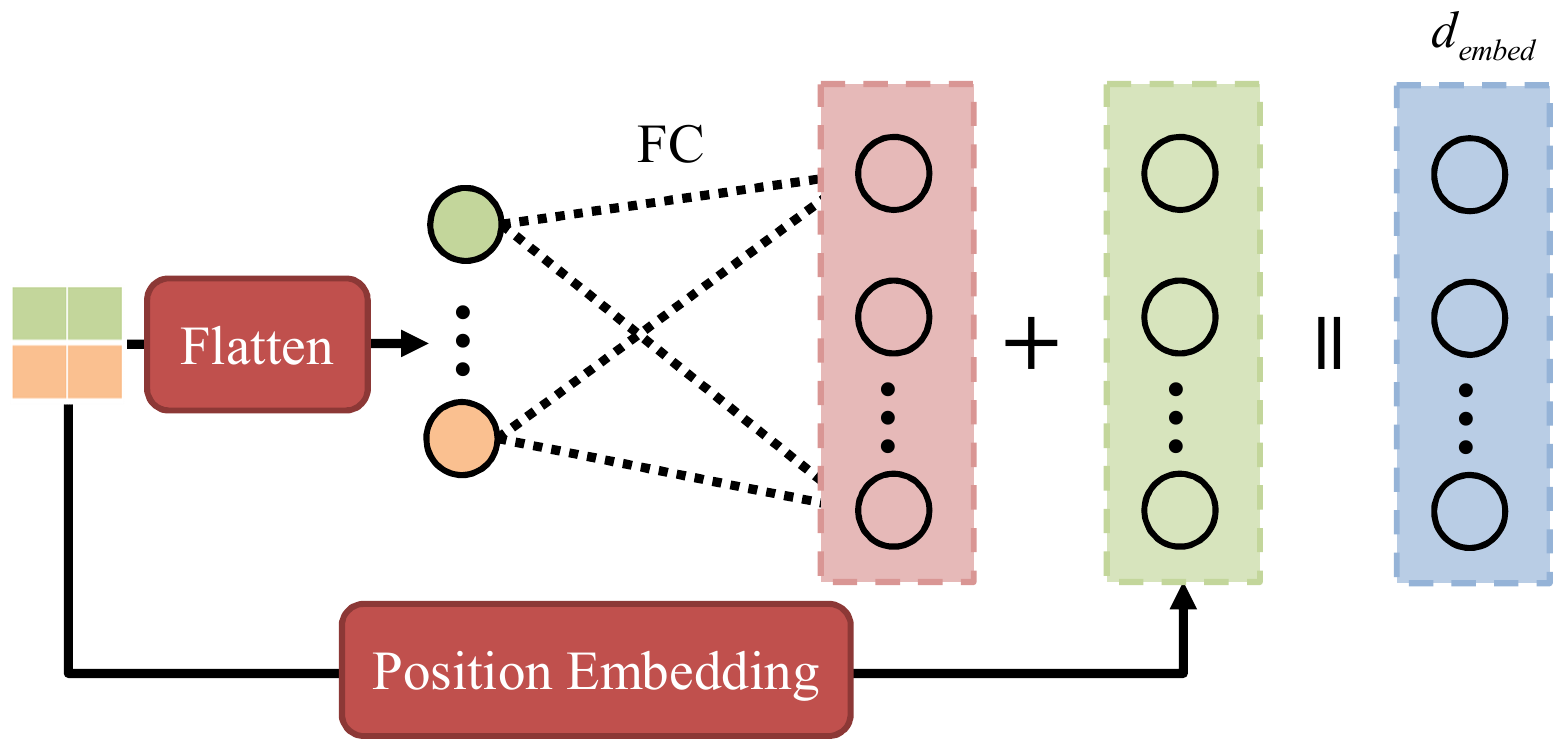}
\caption{Embedding process of battery data patch.}
\label{fig3}
\end{figure}

Each patch contains ${{S}_{patch}}\times {{F}_{patch}}$ data elements. After flattening, it passes through an FC layer and outputs the patch embedding with dimension ${{d}_{embed}}$. Since the Transformer inputs the patch embedding of all positions into the network calculation at the same time, the sequence order will be lost. And in battery data, the order of charging time is very important, so position embedding is needed to solve this problem. The position embedding used by ViT is one-dimensional embedding that can be learned, and the dimension of position embedding is also ${{d}_{embed}}$. Add with the patch embedding obtained before to obtain the input sequence of Transformer encoder, as shown in Fig. \ref{fig3}.

\subsubsection{Transformer Encoder}
Transformer Encoder is mainly divided into two parts, one is the Multi-Head Attention (MHA) layer and the other is Multi-Layer Perceptron (MLP) layer.

\begin{figure}[!t]
\centering
\subfloat[]{\includegraphics[width=1.9in]{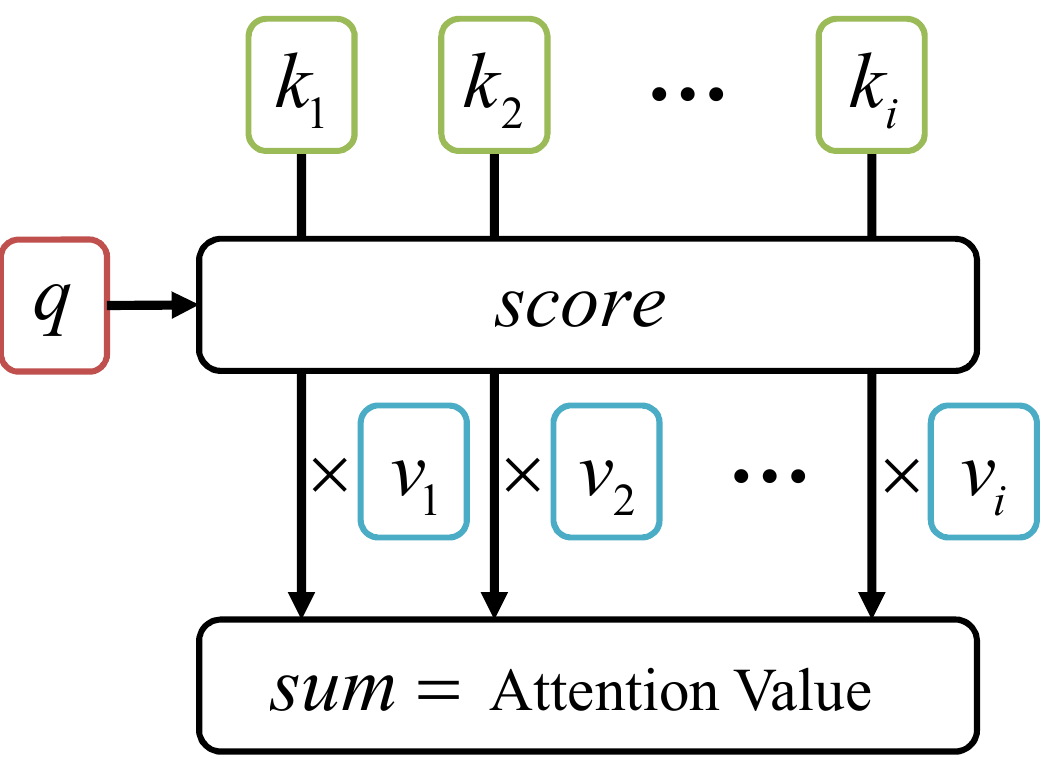}%
\label{fig4a}}
\subfloat[]{\includegraphics[width=1.55in]{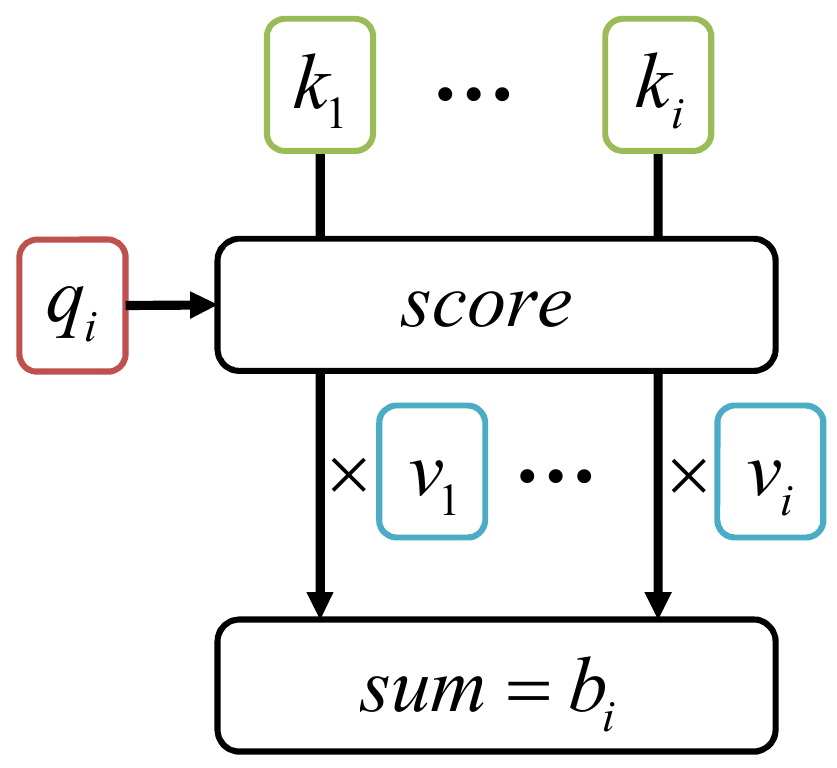}%
\label{fig4b}}
\caption{(a) The calculation process of attention weight. (b) The calculation process of Self-attention.}
\label{fig4}
\end{figure}

The attention mechanism is a series of $\left\langle Key, Value \right\rangle$ data pairs formed by the elements of the input data. At this time, given an element $Query$ in the target task, the weight coefficient is obtained by calculating the correlation between $Query$ and each $Key$, and then $Value$ is weighted and summed to obtain the final attention value. As shown in Fig. \ref{fig4a}. The formula is as follows:
\begin{equation}
\label{eq_3}
atten\left( q,k,v \right)=\sum\limits_{i=1}^{N}{score\left( q,{{k}_{i}} \right)\times {{v}_{i}}}
\end{equation}
where $N$ is the number of data pairs of input data, ${{k}_{i}},{{v}_{i}}$ is the $i$-th input data pair, and $score$ is the weight calculation function, mainly including additive model, multiplication model, dot product model, etc.

The basis of MHA is the Self-attention mechanism. Compared with other attention methods for finding associated weight between $Query$ in the target task and $Key$ in the input data, Self-attention is from the input data, and weight coefficients can be calculated between the input data, as shown in Fig. \ref{fig4b}. The calculation formula is as follows:
\begin{equation}
\label{eq_4}
{{q}_{i}}={{w}_{q,i}}{{a}_{i}}
\end{equation}
\begin{equation}
\label{eq_5}
{{k}_{i}}={{w}_{k,i}}{{a}_{i}}
\end{equation}
\begin{equation}
\label{eq_6}
{{v}_{i}}={{w}_{v,i}}{{a}_{i}}
\end{equation}
where ${{a}_{i}}$ is the $i$-th input of the Self-attention mechanism. ${{w}_{q,i}}$ is used to calculate the $Query$ ${{q}_{i}}$ corresponding to input ${{a}_{i}}$. ${{w}_{k,i}}$ is used to calculate the $Key$ ${{k}_{i}}$ corresponding to input ${{a}_{i}}$. ${{w}_{v,i}}$ is used to calculate the $Value$ ${{v}_{i}}$ corresponding to input ${{a}_{i}}$. Finally, the output ${{b}_{i}}$ corresponding to the input ${{a}_{i}}$ is obtained by adding.

In this paper, the associated weight is calculated by using the scaled dot product model. Therefore, matrix calculation can be used to calculate the weight at the same time to simplify the score calculation operation. The scaled dot product model adds a scaled denominator to the dot product model. This is mainly because when the absolute value of the input matrix is too large, the gradient of the softmax function will become too small to affect the gradient drop. The formula is as follows:
\begin{equation}
\label{eq_7}
Atten\left( Q,K,V \right)=softmax\left( \frac{Q{{K}^{T}}}{\sqrt{{{d}_{atten}}}} \right)V
\end{equation}
where $Q=\left[{{q}_{1}},\ldots,{{q}_{i}}\right],K=\left[{{k}_{1}},\ldots,{{k}_{i}}\right],V=\left[{{v}_{1}},\ldots,{{v}_{i}}\right]$ are the matrices spliced by $Query$, $Key$ and $Value$ respectively. And the softmax function is used to get the attention weight matrix. ${{d}_{atten}}$ is the dimension of $q,k,v$. 

The scaled dot product model is used to calculate the Self-attention weight, which can simplify the calculation difficulty. But it also leads to no learning parameters in the process of weight calculation. Therefore, $Q, K$ and $V$ are linearly projected into multiple groups of $\overline{Q},\overline{K}$ and $\overline{V}$, and the parameters of the projection process can be trained. Each $\overline{Q},\overline{K}$ and $\overline{V}$ focuses on different parts of the input information and then splices as shown in Fig. \ref{fig5}. The formula is as follows:
\begin{figure}[!t]
\centering
\includegraphics[width=2.5in]{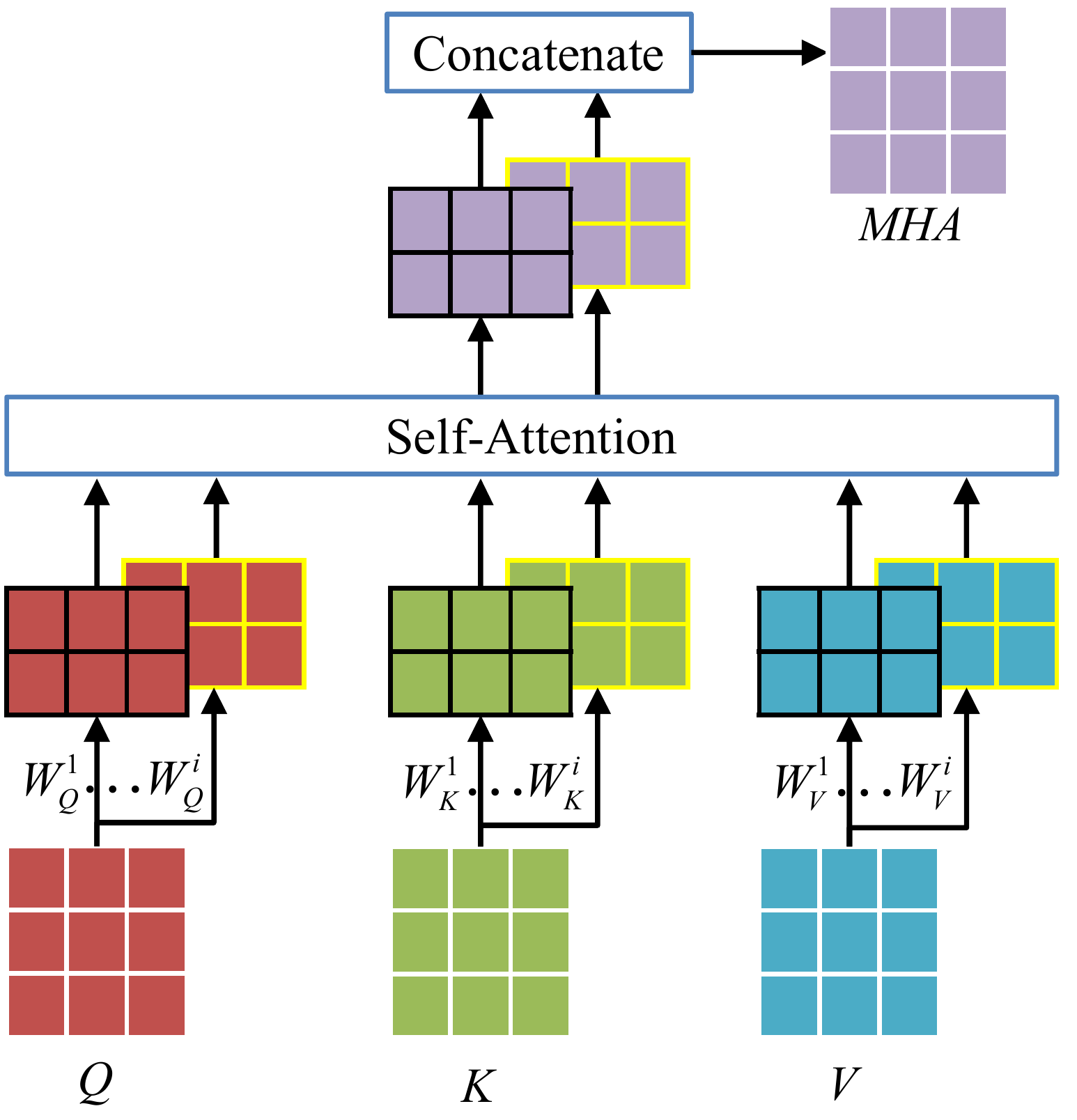}
\caption{The calculation process of Multi-Head Attention.}
\label{fig5}
\end{figure}
\begin{equation}
\begin{aligned}
\label{eq_8}
MHA\left( Q,K,V \right)&=\\
Atten&\left( QW_{Q}^{1},KW_{K}^{1},VW_{V}^{1} \right)\oplus \cdots \\
&\oplus Atten\left( QW_{Q}^{h},KW_{K}^{h},VW_{V}^{h} \right)
\end{aligned}
\end{equation}
where $\oplus$ denotes the splicing of the matrix. $W_{Q}^{h}$ is the linear projection parameter matrix of matrix $Q$ in the $h$-th projection mode. $W_{K}^{h}$ is the linear projection parameter matrix of matrix $K$ in the $h$-th projection mode. $W_{V}^{h}$ is the linear projection parameter matrix of matrix $V$ in the $h$-th projection mode. And $W_{Q}^{h},W_{K}^{h},W_{V}^{h}\in {{\mathbb{R}}^{{{d}_{atten}}\times {{d}_{head}}}}$, ${{d}_{head}}$ is the vector dimension after projection. To avoid the additional calculation cost of the scaling dot product process, this paper takes ${{d}_{head}}={{d}_{atten}}/h$.

Finally, the output of Transformer Encoder is obtained through a two-layer MLP. MLP has two main functions. One is dimension transformation, which ensures that the input dimension of Transformer Encoder is the same as the output dimension. The other is to introduce a nonlinear activation function to make up for the disadvantage that MHA only uses linear projection to obtain stronger model expression ability.

\begin{figure}[!t]
\centering
\includegraphics[width=3in]{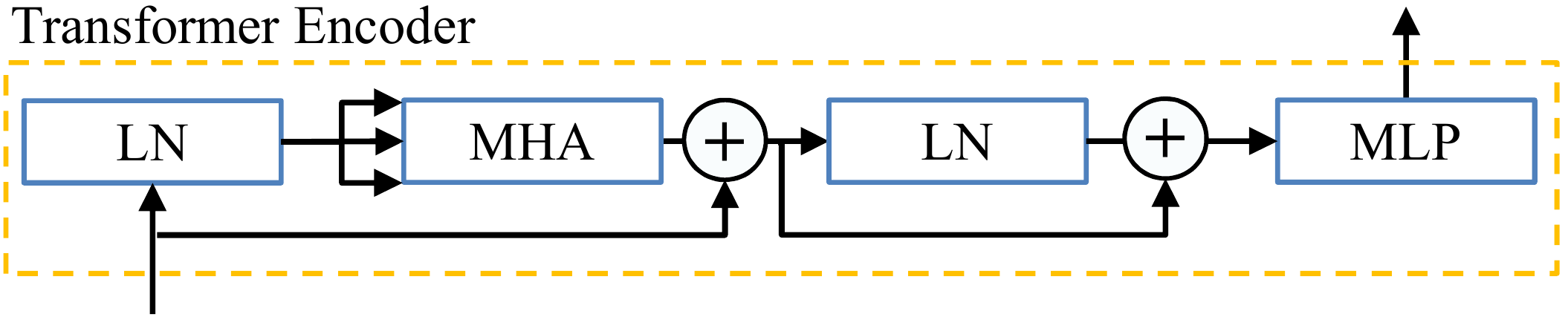}
\caption{The structure of Transformer Encoder.}
\label{fig6}
\end{figure}

In addition, the residual connections \cite{2017Residual} are used before MLP and MHA, which is conducive to increasing the depth of the network. Layer normalization (LN) \cite{2016Layer} can improve the training speed and accuracy of the model and make the model more robust. The input and output dimensions of the encoder layer are the same, so the superposition of multiple transformer encoders constitutes the feature extraction module of ViT-FC as shown in Fig. \ref{fig6}.

\subsubsection{FC for regression}
\begin{figure}[!t]
\centering
\includegraphics[width=3in]{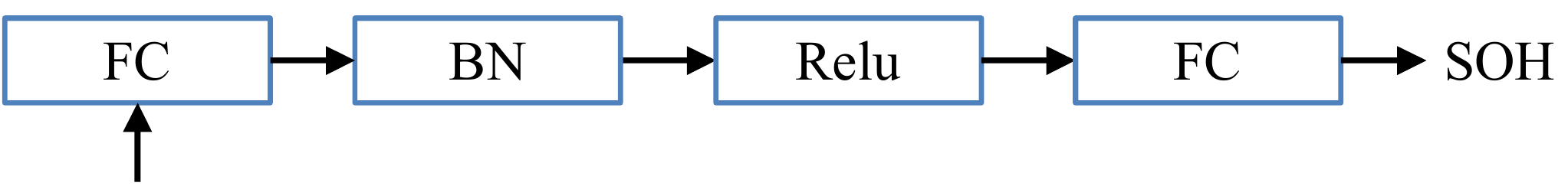}
\caption{The structure of the Full Connection layer.}
\label{fig7}
\end{figure}

Finally, an FC layer, as shown in Fig. \ref{fig7}, is used to establish the dependency between the advanced features obtained by the ViT and the corresponding SOH. Batch normalization (BN) \cite{2015Batch} is used to reduce gradient disappearance and accelerate the convergence process. Relu activation function is used to provide nonlinear features and learn complex relationships of data.

\subsection{Transfer learning}
In some deep learning scenarios, the cost of training the target data directly from scratch is too high, so researchers expect to use some existing relevant knowledge to assist the learning of new knowledge. Transfer learning is a method to solve this problem by transferring the learned model parameters to help the training of the new model. The core of transfer learning is to find the correlation between existing knowledge and new knowledge.

In the SOH prediction problem, the research object is the SOH corresponding to different cycle conditions, and the correlation is high. The knowledge learned in the source task battery can be transferred to the prediction task of the target task battery.

The function of the training set of the source task is to fit the model parameters, gradient the training error during the training process, and learn the ageing law of the source task battery. In the training set of the source task, 20\% of the data are randomly selected as the validation set and do not participate in the training of the model. The function of the validation set is to preliminarily evaluate the learning ability of the model during the training process, and stop the training process in advance to reduce the problem of model overfitting. The test set of the source task is used to evaluate the learning and generalization ability of the model on the dataset of the source task. 

The training set of the target task only contains a small amount of data at the beginning of the battery ageing phase of the target task, which is used to fine-tune the trained model in the source task. In the process of fine-tuning, all parameters of the ViT feature extraction layer in the model are frozen, so the feature extraction methods learned in the training process of the source task are retained. Fine-tune and update all parameters of the FC layer of regression prediction. The FC layer acts as a "firewall" in the process of model representation capability transfer \cite{2017In}. The FC layer can also ensure the transfer of the model representation ability when the battery conditions of the target task and the source task differ greatly.

\section{Experiment}
To ensure the prediction effect of the ViT-FC model, this paper first uses grid search to find the best hyperparameters of ViT-FC model. The prediction results of ViT-FC are compared with those of CNN based on cyclic features and LSTM based on ageing features. Then, the sensitivity of CNN and ViT-FC to the type, scale and discretization granularity of input data is tested. Finally, the transfer ability of different models was verified by two unknown batteries. The experiment runs on Intel Core Processor i5-12600 CPU (3.2 GHz), NVIDIA RTX 3080 GPU with 12 GB GDDR6X and 64 GByte RAM.
\subsection{Definition of error measures}
To evaluate the SOH prediction effect of models, Root Mean Square Percentage Error (RMSPE), Mean Absolute Percentage Error (MAPE) and Standard Deviation of the Error (SDE) are used as error measures in this paper. The formula is as follows:
\begin{equation}
\label{eq_9}
RMSPE=\sqrt{\frac{1}{m}\sum\limits_{i=1}^{m}{{{\left( \frac{{{y}_{i}}-{{{\hat{y}}}_{i}}}{{{y}_{i}}} \right)}^{2}}}}\times 100\%
\end{equation}
\begin{equation}
\label{eq_10}
MAPE=\frac{1}{m}\sum\limits_{i=1}^{m}{\left| \frac{{{y}_{i}}-{{{\hat{y}}}_{i}}}{{{y}_{i}}} \right|\times 100}\%
\end{equation}
\begin{equation}
\label{eq_11}
SDE=\sqrt{\frac{1}{m}{{\sum\limits_{i=1}^{m}{\left( {{x}_{i}}-\bar{x} \right)}}^{2}}}
\end{equation}
where $m$ is the number of data participating in the calculation of error. ${{y}_{i}}$ is the real value of SOH and ${{\hat{y}}_{i}}$ is the predicted value of SOH. ${{x}_{i}}={{y}_{i}}-{{\hat{y}}_{i}}$ is the error value of SOH prediction, and $bar{x}$ is the average value of the error.

\subsection{The configuration of ViT-FC }
In this paper, grid search \cite{YAO2021118866} is used to obtain the optimal model parameters. ViT-FC is trained by exhausting the search parameters within the specified parameter range. And the parameters with the highest accuracy in the validation set are found from all the parameters. This is a process of training and comparison. Finally, the selected parameters are shown in Table \ref{table2}.
\begin{table}[!t]
\caption{The configuration of ViT-FC}
\label{table2}
\centering
\begin{tabular}{|c||c|}
\hline
Hyperparameters & \\
\hline
Learning rate & 0.001\\
Batch size & 16\\
Early stop epoch & 5000\\
Normalization method & MinMax\\
Dropout & 0.1\\
The discretized data included in the patch ${{S}_{patch}}$ & 20\\
the type of battery data included in the patch ${{F}_{patch}}$ & 2\\
The dimension of embedding ${{d}_{embed}}$ & 512\\
The number of attention heads $h$ & 8\\
The dimension of attention head ${{d}_{head}}$ & 64\\
Hidden neurons of MLP in Encoder & 512\\
Hidden neurons of FC & 32\\
\hline
\end{tabular}
\end{table}

\begin{figure}[!t]
\centering
\includegraphics[width=2.7in]{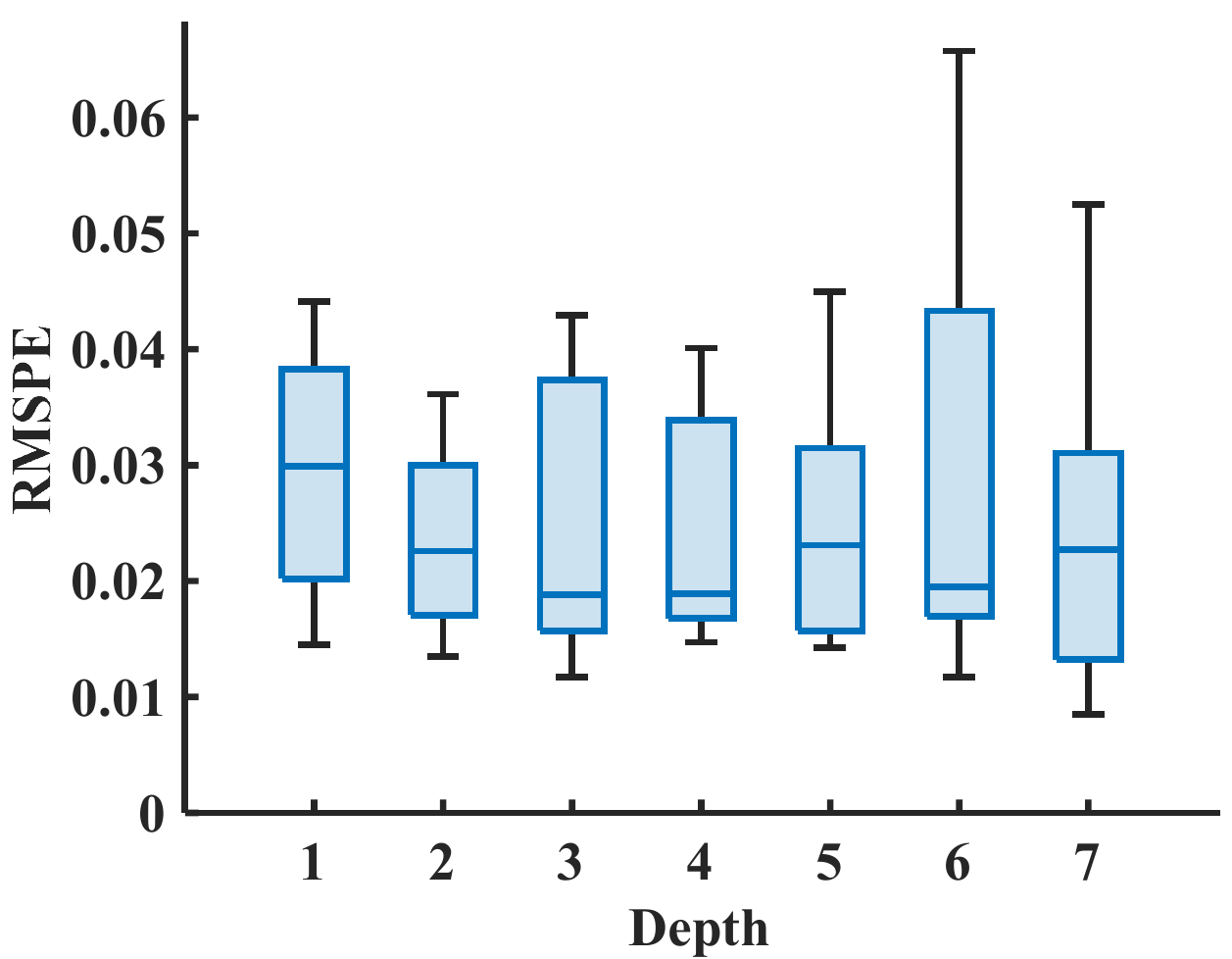}
\caption{Influence of changing network depth on prediction accuracy of ViT-FC.}
\label{fig8}
\end{figure}

Among the parameters of ViT-FC, the number of Transformer Encoders, which is the depth of the network, is highly related to the effectiveness of feature acquisition. To study the influence of network depth on the accuracy of the SOH prediction model, the following experiments were designed. The discretization granularity $S$ is set to 200 and the training set ratio ${{R}_{t}}$ is set to 0.5. Ten groups of randomly divided datasets were used to train the ViT-FC model, and the RMSPE box diagram as shown in Fig \ref{fig8} is obtained.

It can be seen from the figure that the deeper the network, the larger the capacity of the network and the stronger the feature expression ability. In the first three layers, each additional layer can bring good effect improvement. When the depth of the network is continued to increase, the prediction effect does not improve significantly and even deteriorates significantly at 6 layers. Finally, considering the effect of prediction and the speed of training, the depth of the network is set as 4.

\subsection{Baseline}
According to the existing SOH prediction methods, the CNN method based on cycle features and the LSTM method based on ageing features are selected as the baseline methods.

CNN: The deep learning method for online capacity prediction of LIBs was initially introduced \cite{SHEN2019100817}. A two-layer CNN structure is designed using a similar structure in the paper. The configuration of CNN is shown in Table \ref{table3}.
\begin{table}[!t]
\caption{The configuration of CNN}
\label{table3}
\centering
\begin{tabular}{|c||c|}
\hline
Hyperparameters & \\
\hline
Learning rate & 0.001\\
Batch size & 16\\
Early stop epoch & 5000\\
Normalization method & MinMax\\
Dropout & 0.1\\
The number of convolution kernels of Conv. 1 & 6\\
The size of convolution kernels of Conv. 1 & [5,2]\\
The stride of convolution kernels of Conv. 1 & 1\\
The size of pooling. 1 & [2,2]\\
The stride of pooling. 1 & [2,1]\\
The number of convolution kernels of Conv. 2 & 16\\
The size of convolution kernels of Conv. 2 & [5,1]\\
The stride of convolution kernels of Conv. 2 & 1\\
The size of pooling. 2 & [2,1]\\
The stride of pooling. 2 & [2,1]\\
Hidden neurons of FC & 32\\
\hline
\end{tabular}
\end{table}

LSTM: it has advantages in capturing long-term correlation and is one of the popular methods in the field of SOH and remaining life prediction in recent years. Due to the weak ability of LSTM to capture battery cycle features, the charging time of ${{V}_{low}}$ to ${{V}_{high}}$ is selected as the input feature in this paper \cite{2021xing}. The number of LSTM layers is set to 5, the hidden neurons of the LSTM layer are set to 256, and the hidden neurons of FC are set to 32.

\subsection{The predicted result of the source task}
Only the current, voltage and temperature data of the battery are used. The ratio of the training set ${{R}_{t}}$ is set to 0.7 and the discretization granularity $S$ is set to 100. The results are shown in Table \ref{table4}. The effect of ViT-FC is good under all indexes.
\begin{table}[!t]
\caption{Training results of source task}
\label{table4}
\centering
\begin{tabular}{|c||c c c|}
\hline
Method & ViT-FC & CNN & LSTM\\
\hline
RMSPE & \textbf{0.323\%} & 0.614\% & 1.586\%\\
MAPE & \textbf{0.165\%} &0.241\% & 1.049\%\\
SDE & \textbf{1.112\%} & 2.070\% & 4.299\%\\
\hline
\end{tabular}
\end{table}

\begin{figure}[!t]
\centering
\subfloat[]{\includegraphics[width=1.7in]{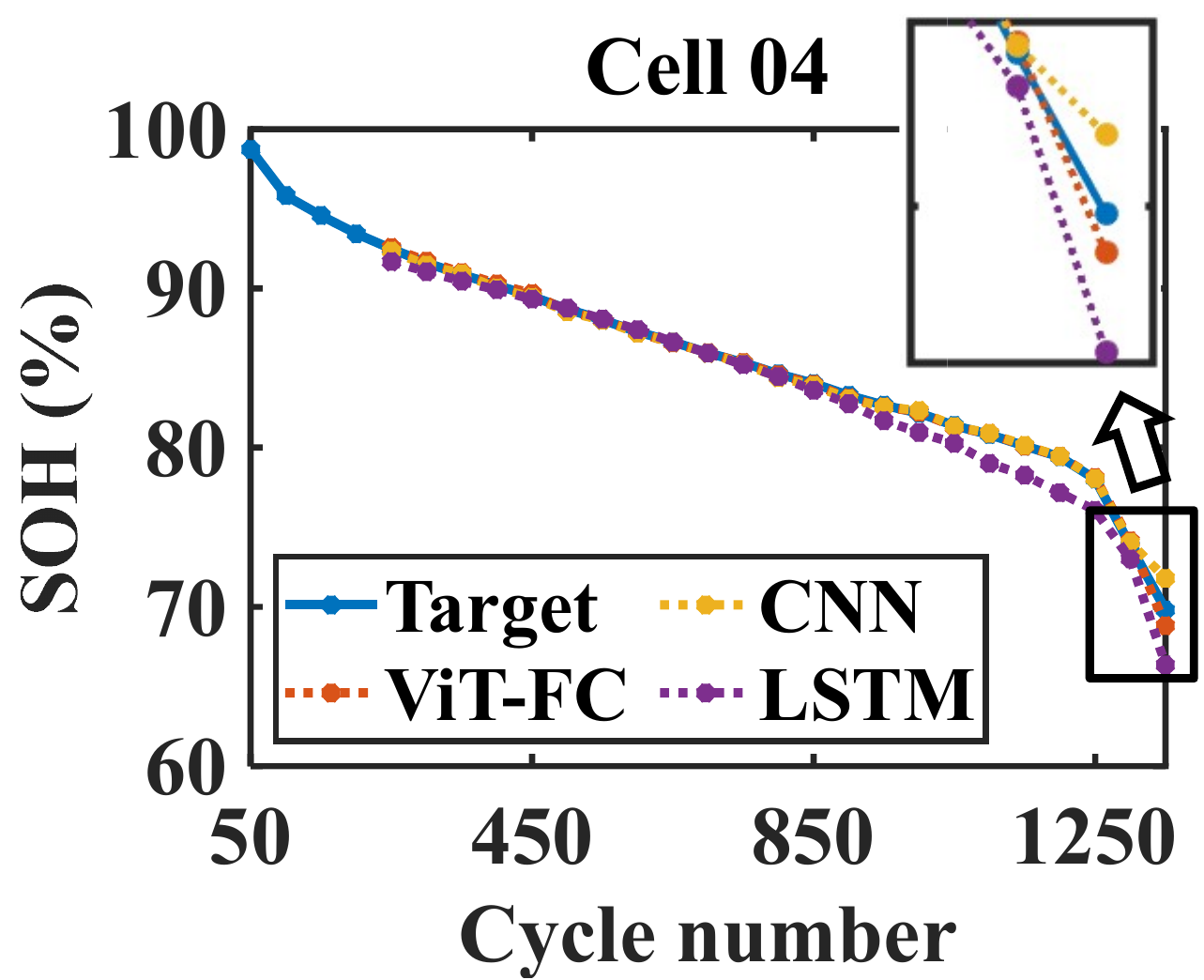}%
\label{fig9a}}
\hfil
\subfloat[]{\includegraphics[width=1.7in]{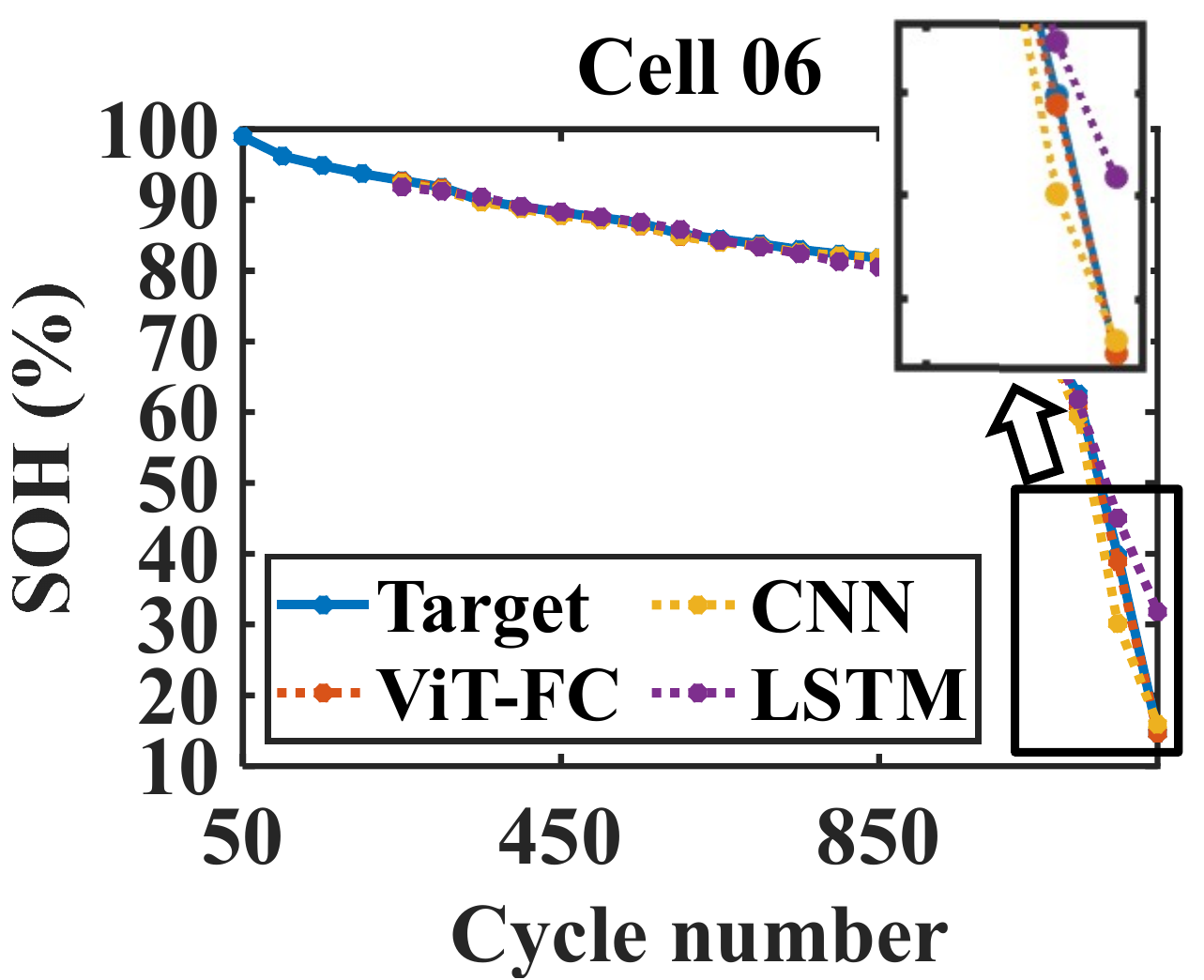}%
\label{fig9b}}
\caption{(a) The prediction results of Cell 04. (b) The prediction results of Cell 06.}
\label{fig9}
\end{figure}

Fig \ref{fig9} shows the prediction results of Cell 04 and Cell 06. It can be observed from the figure that all models have a good prediction effect in 80\% to 90\% SOH segment. Under 80\% SOH, the prediction effect of LSTM based on ageing features is slightly worse than CNN and ViT-FC based on cyclic features.

\begin{figure}[!t]
\centering
\subfloat[]{\includegraphics[width=1.7in]{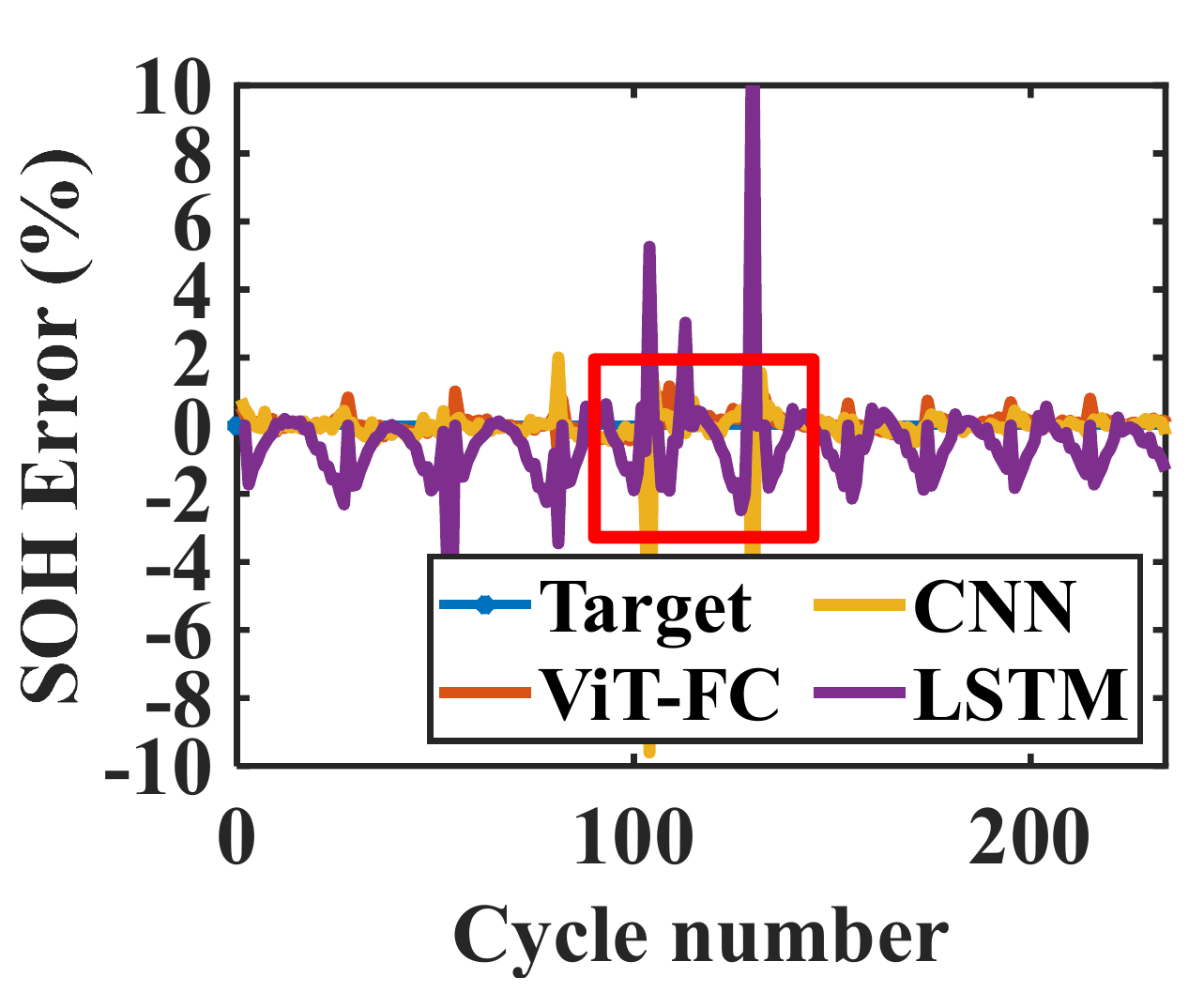}%
\label{fig10a}}
\hfil
\subfloat[]{\includegraphics[width=1.7in]{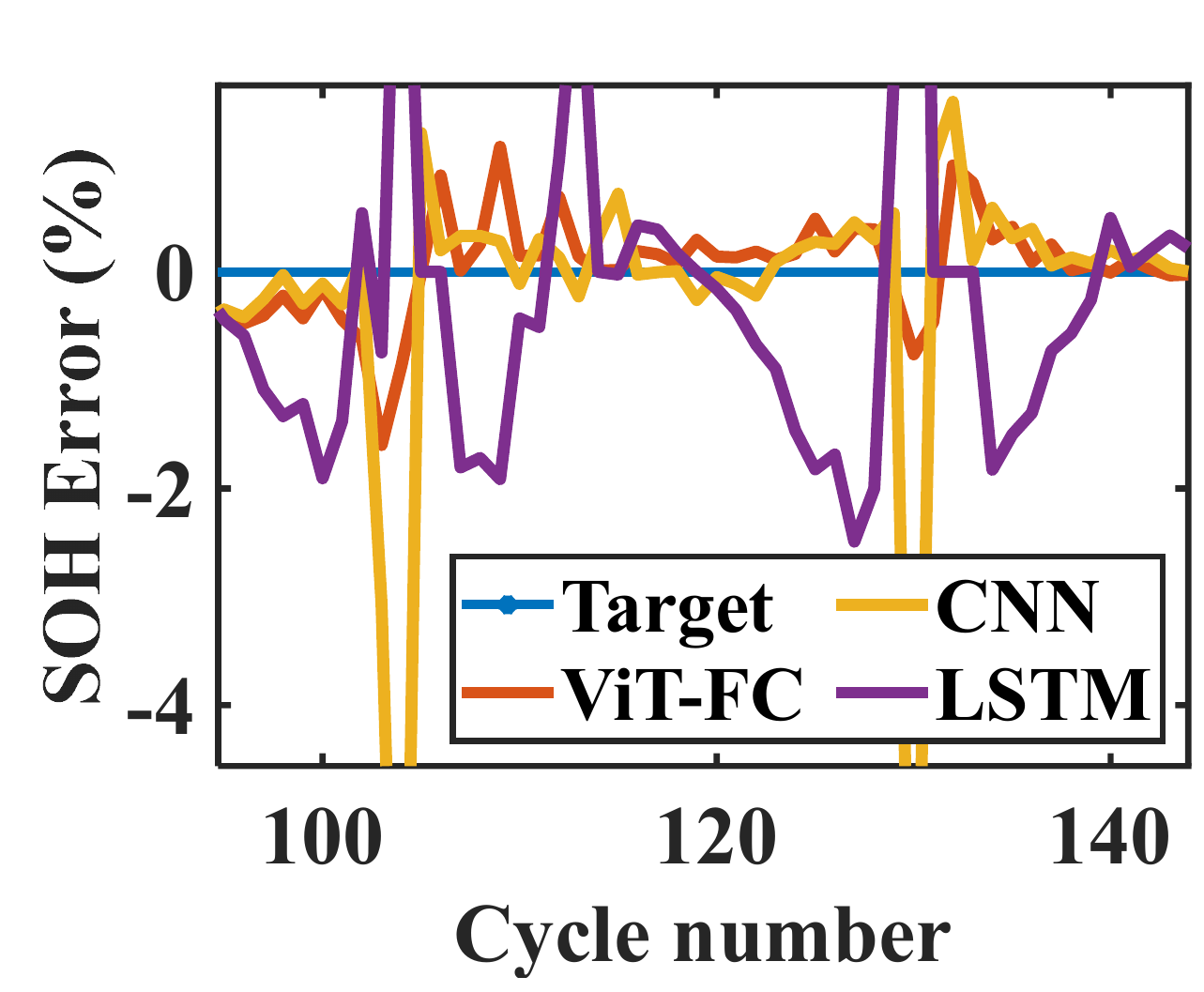}%
\label{fig10b}}
\caption{(a) The prediction error figure of the source task battery. (b) Enlarged view of the red box in Fig. \ref{fig10a}}
\label{fig10}
\end{figure}

Fig \ref{fig10} is the prediction error figure of the source task battery. It can be seen from the figure that the prediction result of LSTM is generally low, with an error of 2\% or more at the beginning and end of the battery life cycle. CNN and ViT-FC can keep the SOH error within 1\% in most cases. However, the error of CNN will be large at the end of the battery life cycle. Especially under the extreme working condition, it even reaches 8\%. At the same time, under the same extreme working condition and the same ageing degree, the error of ViT-FC still does not exceed 2\%.

\subsection{The sensitivity to the type of input data}
During the operation of the battery, the current, voltage and temperature of the battery can be directly measured by the sensor. According to the electrochemical characteristics of the battery, some battery data that cannot be directly measured, such as internal resistance and capacity, can be obtained from the directly measured battery data. To study the sensitivity of the proposed method to the type of input data, the Hybrid Pulse Power Characterization (HPPC) is used to obtain the internal resistance of the battery. HPPC is based on IEC62660-1 standard \cite{1997Secondary}. During the test, the applied HPPC current pulse can obtain the voltage response of the battery, and the internal resistance of the battery can be obtained by the following formula:
\begin{equation}
\label{eq_12}
{{R}_{in}}=\frac{\Delta V}{\Delta I}
\end{equation}
where $\Delta V$ represents the voltage change at the end of the charge/discharge pulse. $\Delta I$ represents the current pulse.

The coulomb counting method is to calculate the charge and discharge of LIBs by calculating the integral of the current and time and then comparing it with the rated power of the battery to obtain the current remaining power. And the formula is as follows:

\begin{equation}
\label{eq_13}
C={{C}_{N}}-\int\limits_{0}^{t}{\eta Id\tau }
\end{equation}
where ${{C}_{N}}$ is the rated capacity of the battery, $I$ is the current of the battery, and $\eta $ is the charge and discharge efficiency.

\begin{figure}[!t]
\centering
\includegraphics[width=3in]{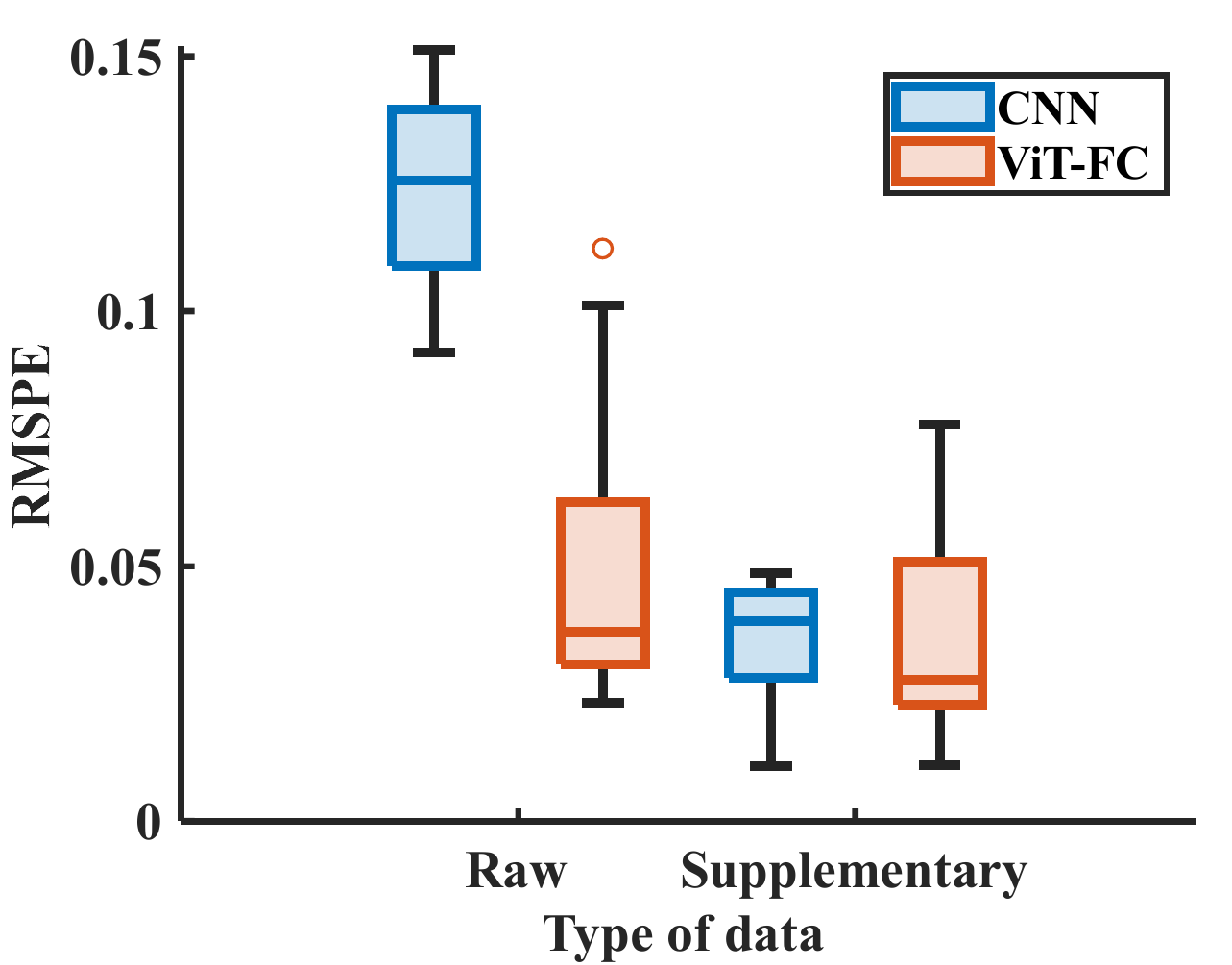}
\caption{Influence of the type of input data.}
\label{fig11}
\end{figure}

A control experiment is designed in this paper. One group of experiments only uses the current, voltage and temperature directly measured by the sensor as the input, and is named the raw group. The other group of experiments uses the calculated capacity and internal resistance in addition to the directly measured data and is named the supplementary group. The ratio of the training set ${{R}_{t}}$ is set to 0.7 and the discretization granularity $S$ is set to 100. The same ten groups of random datasets are used to train CNN and ViT-FC to obtain the RMSPE box diagram as shown in Fig. \ref{fig11}.

It can be seen that when using CNN, the type of input data directly affects the final training effect. After adding two parameters, capacity and internal resistance, which have a strong relationship with SOH, the prediction effect of CNN have been improved qualitatively. When ViT-FC is used, the type of input data has a relatively small impact on the prediction result. The median and upper and lower quartiles of the predicted RMSPE of ViT-FC decreased, but the overall prediction effect was still similar. When data with increased capacity and internal resistance are used, the median and lower quartile of ViT-FC are still better than CNN, but the convergence of ViT-FC is slightly worse than CNN. Therefore, when only the raw data that can be directly measured by the sensor is used, the ViT-FC can capture more advanced features related to SOH, and the prediction effect is better. After adding the internal parameters of the battery obtained by electrochemical characteristics, the effect of ViT-FC can also reach a good level.

\subsection{The sensitivity to discretization granularity of input data}
\begin{figure}[!t]
\centering
\includegraphics[width=3in]{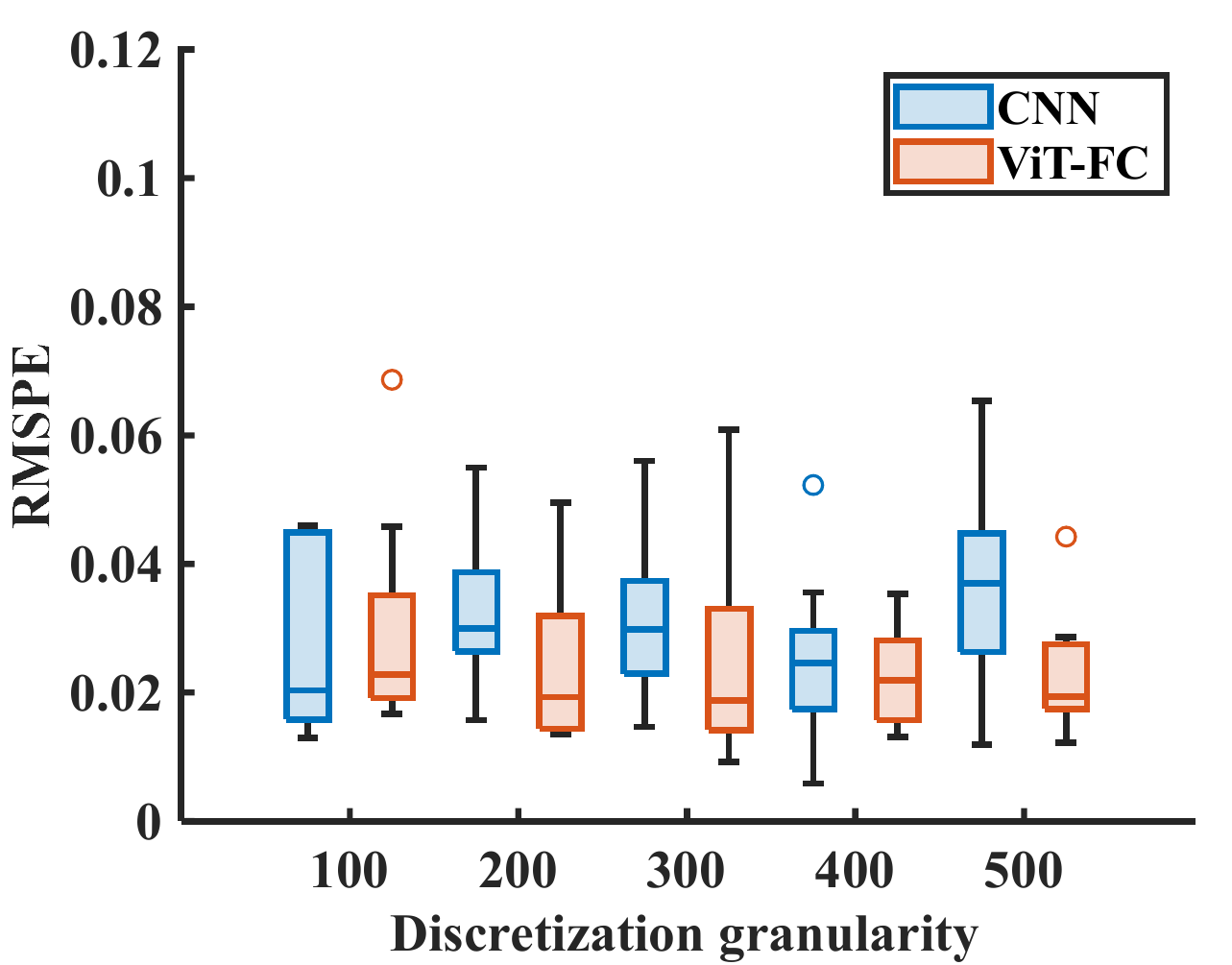}
\caption{Influence of discretization granularity of input data.}
\label{fig12}
\end{figure}

The discretization granularity of input data determines the input matrix size of CNN and ViT-FC. To study the sensitivity of the proposed method to discretization granularity, the discretization granularity $S$ is set to $[100,200,300,400,500]$ and divided into five experimental groups. In each group of experiments, supplementary data were used. The ratio of the training set ${{R}_{t}}$ is set to 0.5. And the source task data set is randomly divided ten times. The RMSPE box diagram of the results is shown in Fig. \ref{fig12}.

It can be seen from the figure that when the discretization granularity is increased from 100 to 200, the median value of RMSPE of ViT-FC is significantly increased. With the continuous increase of discretization granularity, the prediction effect of ViT-FC is generally better, but it deteriorates at 400. However, the median value of RMSPE of CNN has no obvious rule, and when the discretization granularity is 100, it is the lowest. When the discretization granularity is increased, the interquartile range of CNN and ViT-FC is significantly reduced, which has a better convergence effect.

On the whole, the increase of discretization granularity will improve the effect. However, the increase in discretization granularity will lead to the length of embedded sequences and bring greater computational pressure.

\subsection{The sensitivity to the scale of input data}
\begin{figure}[!t]
\centering
\includegraphics[width=3in]{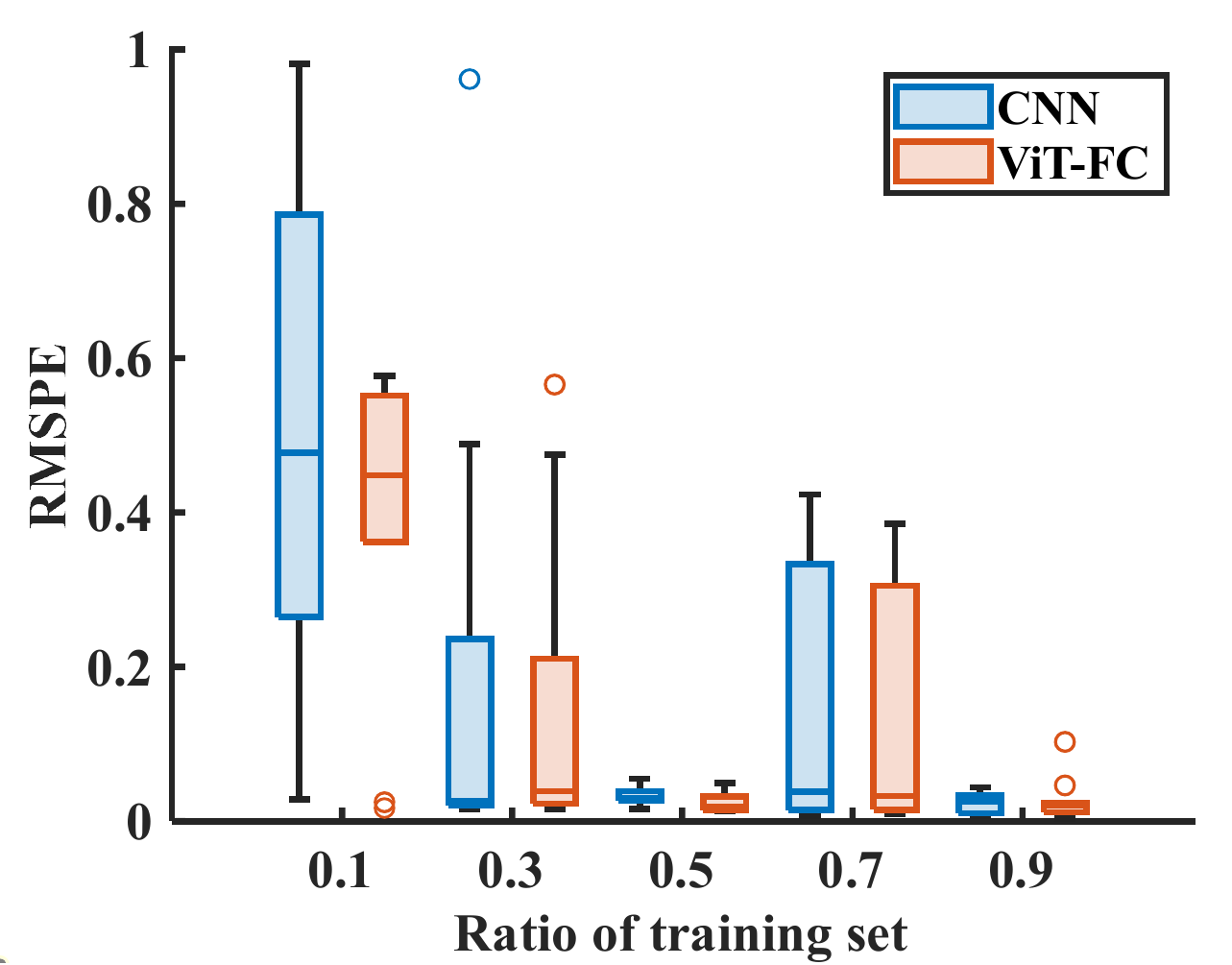}
\caption{Influence of the scale of input data.}
\label{fig13}
\end{figure}

The training process of ViT-FC lacks inductive bias, which makes the model unable to use data efficiently and affects the convergence speed and model performance. Therefore, ViT-FC often requires a large amount of data and a longer training time. To study the sensitivity of the proposed method to the scale of input data, the ratio of the training set ${{R}_{t}}$ is set as $[0.1,0.3,0.5,0.7,0.9]$ and divide into five experimental groups. In each group of experiments, supplementary data were used. the discretization granularity $S$ is set to 200. And the source task data set is randomly divided ten times. The RMSPE box diagram of the results is shown in Fig. \ref{fig13}.

As can be seen from the figure, when ${{R}_{t}}>0.1$, the median value of RMSPE is not significantly optimized with the increase of the training set ratio. But they are better than the RMSPE of ${{R}_{t}}=0.1$. However, it can be seen that when the training set ratio is 0.3 and 0.7, the upper quartile and the maximum value are high, and the stability of the actual application is poor. This is because most of the battery data is within the normal use range, while only a small part of the battery data whose battery life is not up to standard. Then in the process of randomly dividing the dataset, it is very likely that the sample distribution is not uniform.

In general, the convergence and median values of ViT-FC are slightly better than CNN. Even when the amount of data is small, ViT-FC has better performance, which proves that the inductive bias of CNN does not apply to battery data.

\subsection{Transfer learning}
\begin{figure}[!t]
\centering
\includegraphics[width=3in]{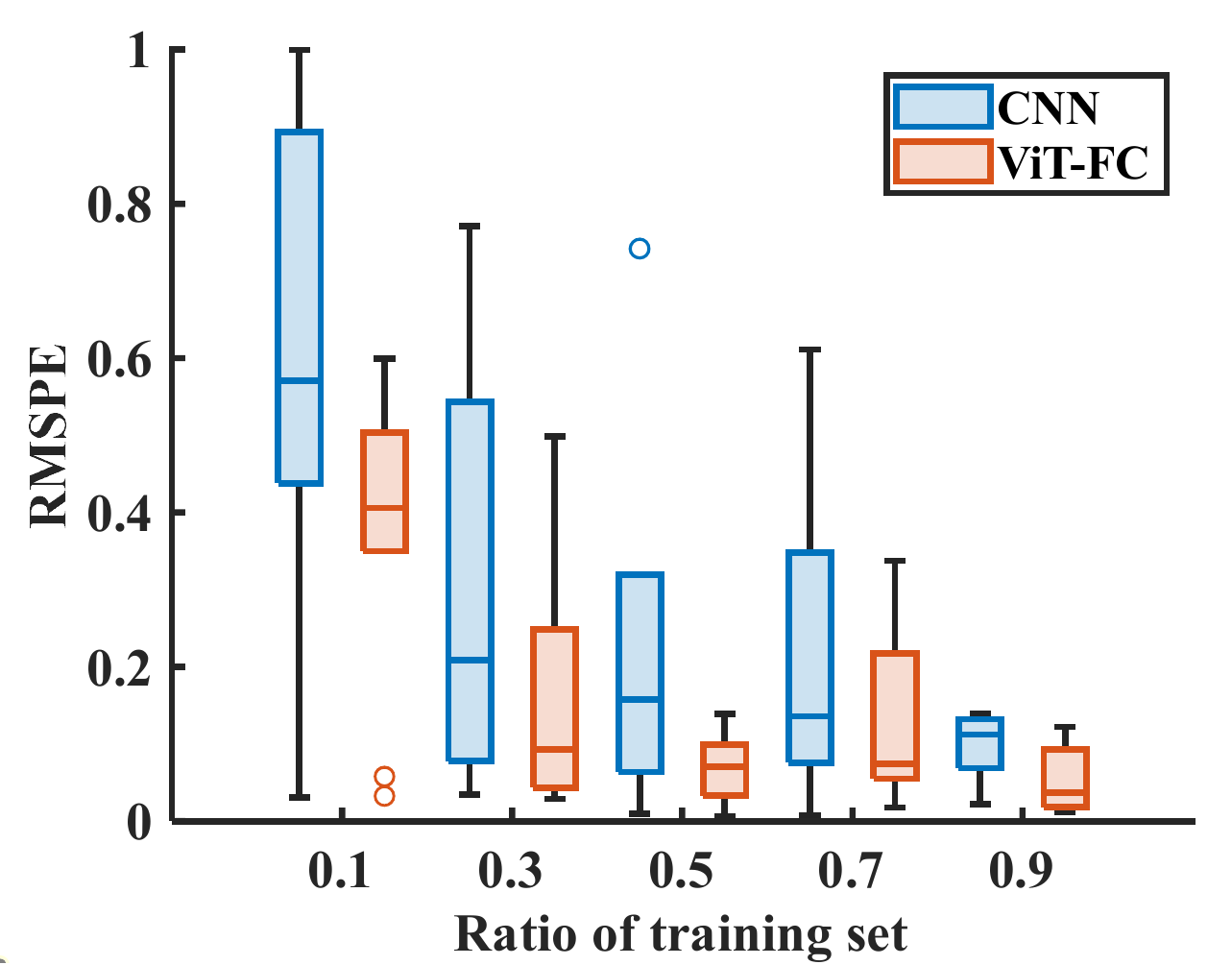}
\caption{Influence of pre-training data size on migration effect.}
\label{fig14}
\end{figure}

In transfer learning, the training set of the target task is only used to fine-tune a few parameters of the regression layer, so the improvement of the transfer effect is limited. The parameters of the feature extraction layer are trained and optimized by the training set of the source task. Therefore, the training set of the source task has a great impact on the transfer effect, as shown in Fig. \ref{fig14}. It can be seen that with the increase of ${{R}_{t}}$, more data are used to train the pre-trained model, and the better the transfer effect is finally obtained.

During the pre-training, supplementary data were used. The ratio of training set ratio is set to 0.9, and the discretization granularity $S$ is set to 200. During fine-tuning, the data of the first 4 cycles of the battery under unknown working conditions are used to optimize the FC layer, and the epoch is set to 20000.
The errors of the target task are shown in Table \ref{table5}. The transfer effect of ViT-FC is significantly ahead of other methods.

\begin{table}[!t]
\caption{The errors of target task}
\label{table5}
\centering
\begin{tabular}{|c||c|c|c|}
\hline
Method & ViT-FC & CNN &  LSTM\\
\hline
\multicolumn{4}{|c|}{Cell 02}\\
\hline
RMSPE&\textbf{0.38\%}&2.28\%&3.71\%\\
MAPE&\textbf{0.33\%}&1.97\%&3.06\%\\
SDE&\textbf{0.85\%}&4.21\%& 8.05\%\\
\hline
\multicolumn{4}{|c|}{Cell 07}\\
\hline
RMSPE&\textbf{2.47\%}&14.53\%&11.17\%\\
MAPE&\textbf{1.02\%}&5.38\%&5.50\%\\
SDE&\textbf{3.45\%}&11.20\%&11.03\%\\
\hline
\end{tabular}
\end{table}

\begin{figure}[!t]
\centering
\subfloat[]{\includegraphics[width=1.7in]{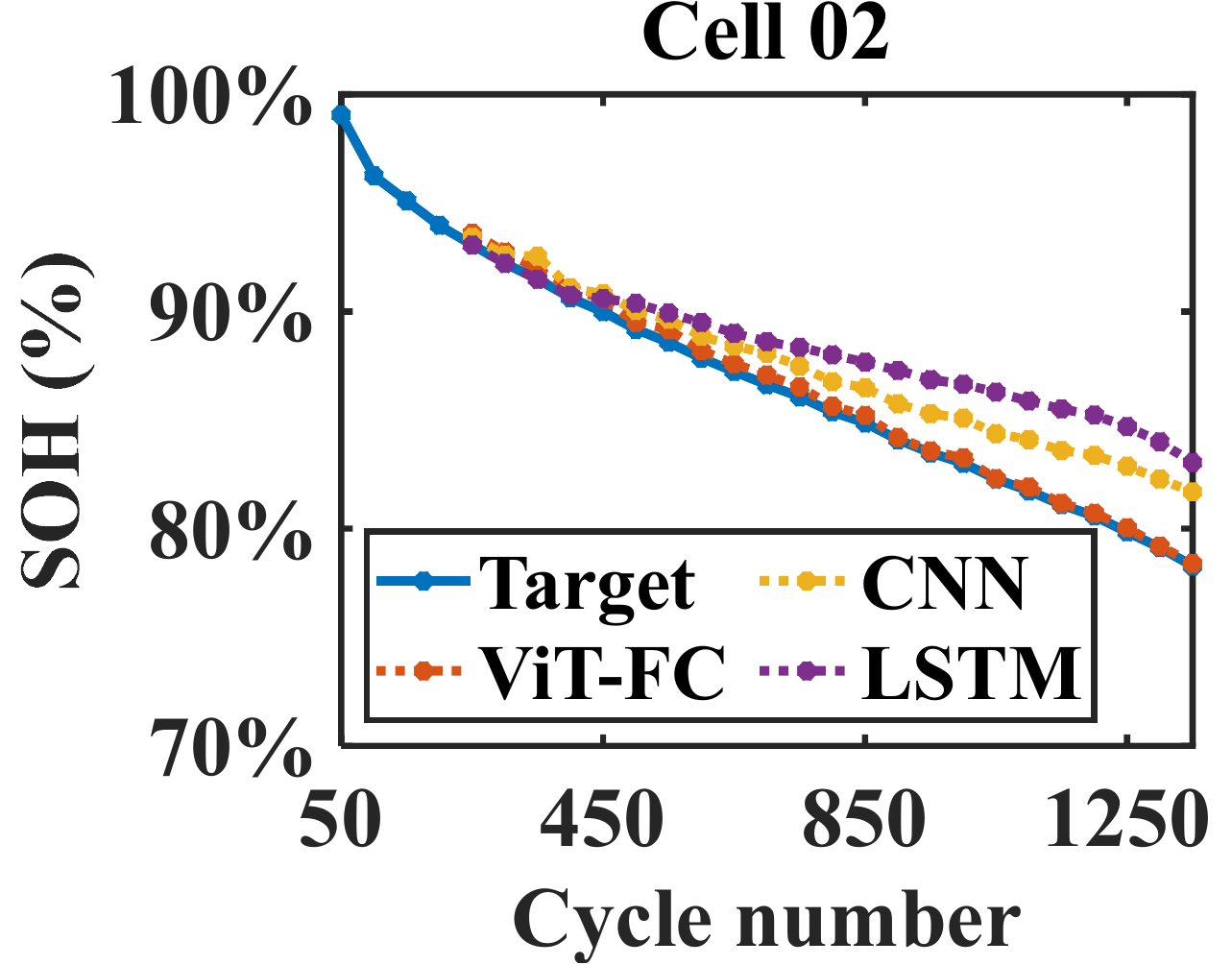}%
\label{fig15a}}
\hfil
\subfloat[]{\includegraphics[width=1.7in]{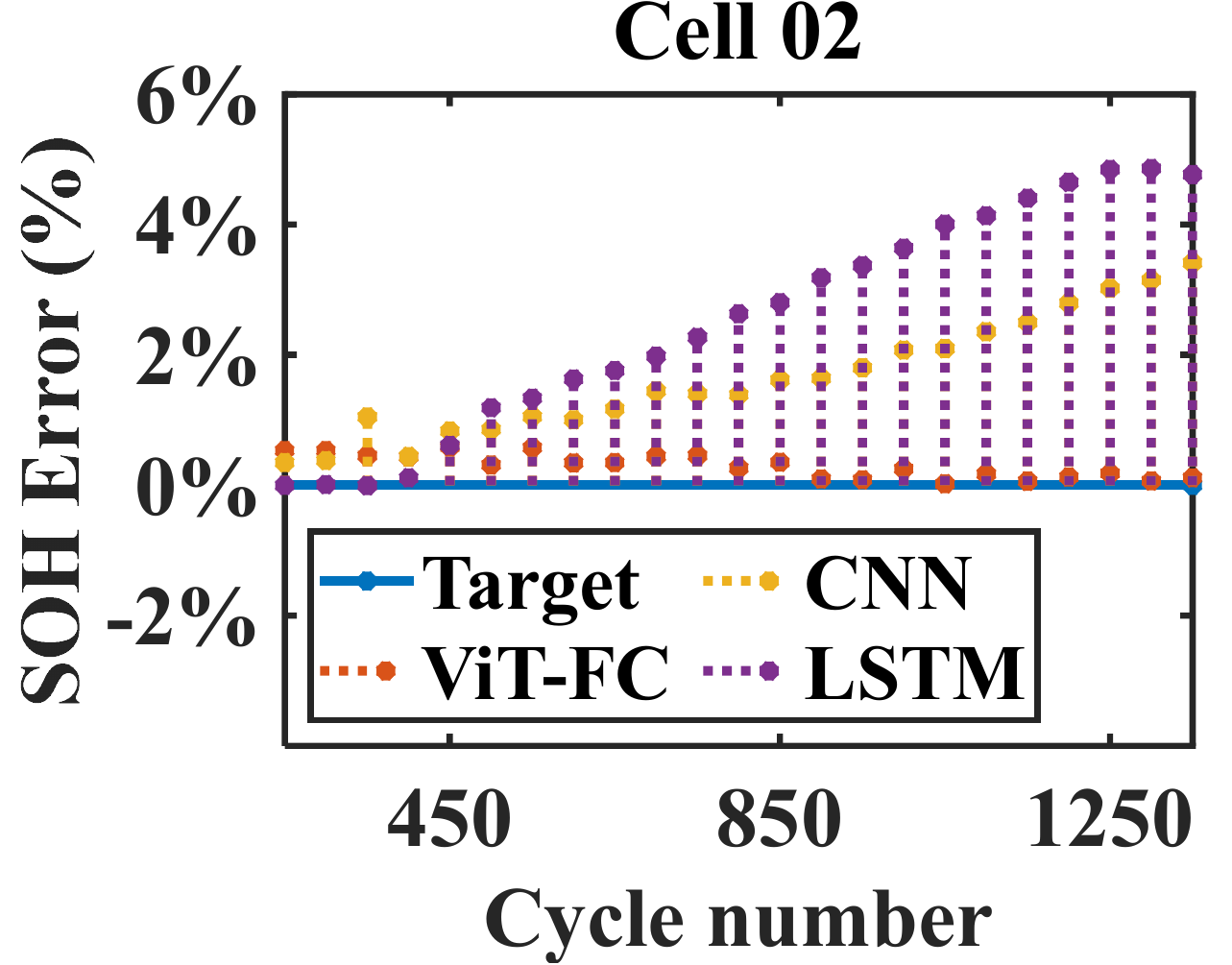}%
\label{fig15b}}
\hfil
\subfloat[]{\includegraphics[width=1.7in]{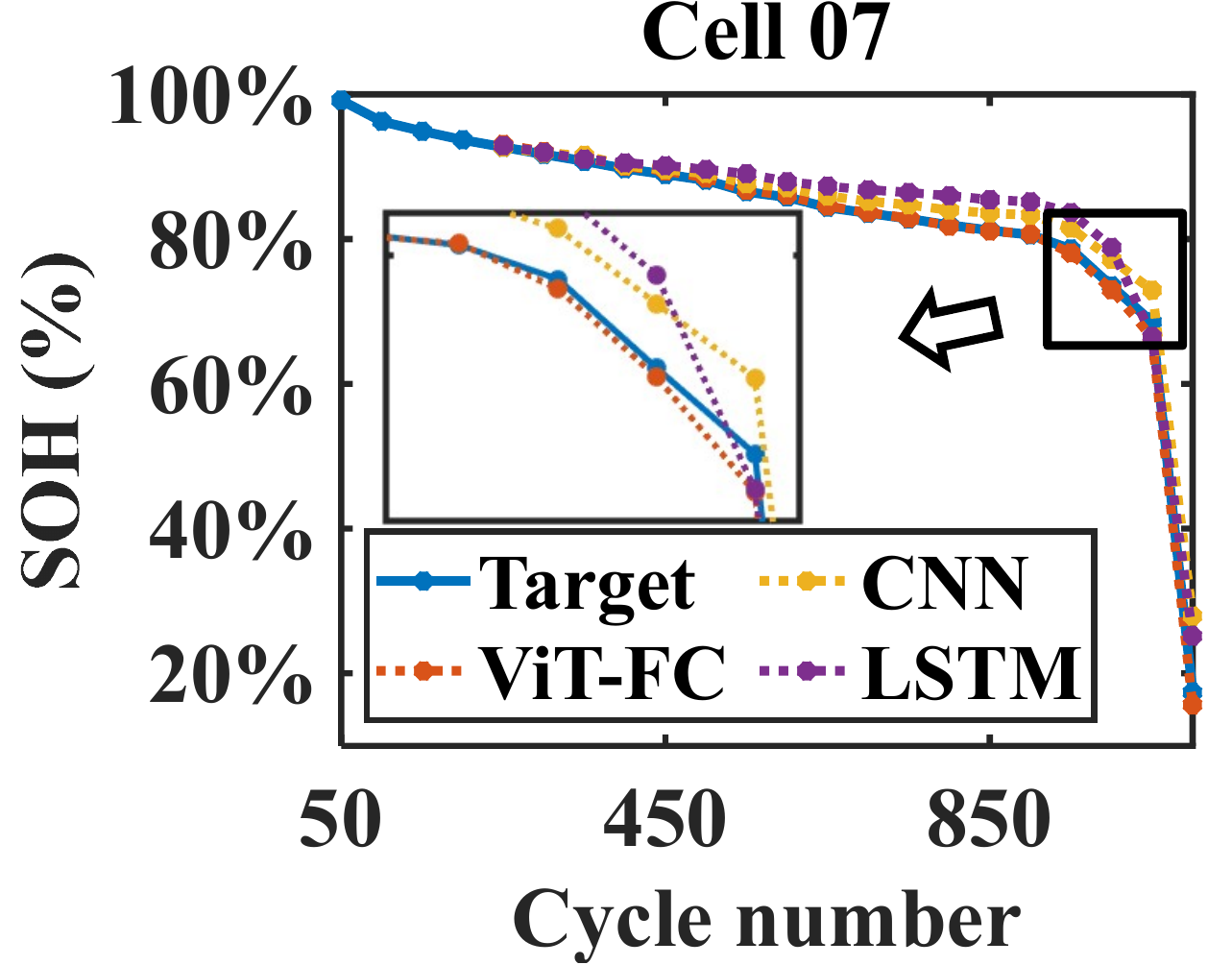}%
\label{fig15c}}
\hfil
\subfloat[]{\includegraphics[width=1.7in]{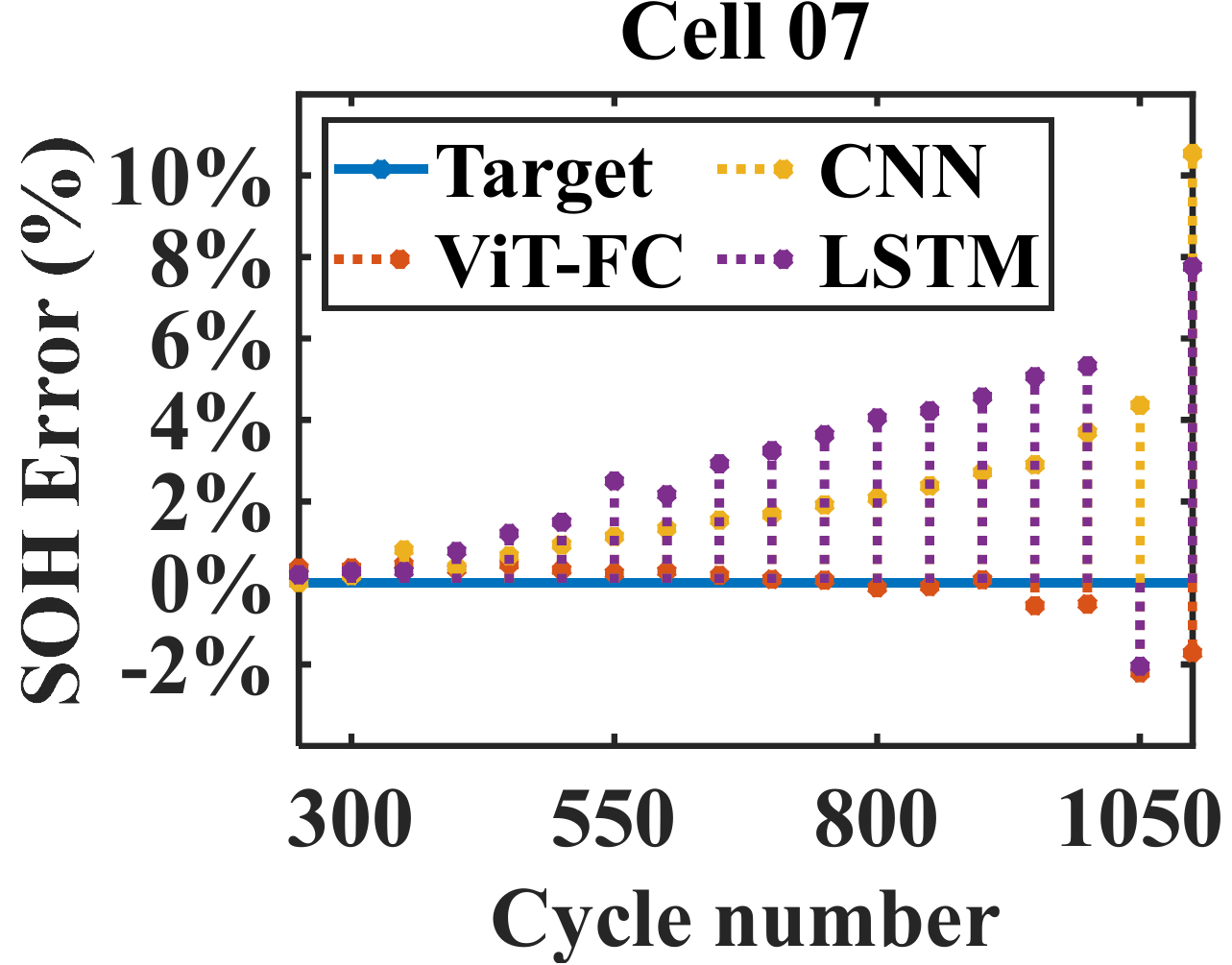}%
\label{fig15d}}
\caption{(a) The prediction results of Cell 02. (b) The prediction error of Cell 02. (c) The prediction results of Cell 07. (d) The prediction error of Cell 07.}
\label{fig15}
\end{figure}

The transfer results of the target task are shown in Fig. \ref{fig15}. It can be seen that all the methods have good prediction results at the beginning of the battery life. With the degradation of the battery, ViT-FC shows a better prediction effect. However, LSTM and CNN have the problem of high predicted SOH, and the deterioration of LSTM is more obvious.  When SOH is greater than 80\%, the transfer effect of cell 07 is similar to that of cell 02. Even if the SOH reaches 20\%, the transfer prediction results of ViT-FC still closely follow the battery ageing trend.

\subsection{Discussion}
The above experimental part verifies the proposed model and the existing SOH prediction model from the aspects of prediction accuracy, data demand, and generalization ability. It can be found that ViT-FC can achieve better prediction accuracy without high-quality data. And a good feature expression of the battery cycle data can be obtained to ensure the transfer effect of the model.

\begin{enumerate}
\item{CNN and ViT-FC based on cycle feature mine the cycle features hidden in the battery charge and discharge cycle data, to prediction SOH. And LSTM based on battery ageing features is to mine the time correlation of ageing through some manually extracted ageing features in historical charge and discharge cycles to predict SOH. The overall effect of the method based on cycle features will be better than the method based on ageing features, and this phenomenon will become more obvious with the increase in battery ageing degree. This is because the method based on ageing features still relies on the ageing features extracted manually, and may not apply to all working conditions.}
\item{CNN and ViT-FC are both based on cycle features, so their prediction results depend on their ability to extract cycle features from the battery charge and discharge cycle data. In general, CNN has the inductive bias of locality and shift invariance, and it will show better performance when training data with small samples. However, the inductive bias of CNN is proposed for the image task and is not applicable to the data matrix of the battery, which makes the inductive bias of CNN become an obstacle in the SOH prediction task.}
\item{When CNN only uses data that can be directly measured, the prediction effect is relatively poor. After adding the internal resistance and capacity obtained by electrochemical characteristics, the prediction effect has been greatly improved. In contrast, the improvement of ViT-FC is not large, and the effect is still better than CNN. This shows that ViT-FC does not need data other than direct measurement data, and can obtain good prediction results. At the same time, ViT-FC only needs part of the measurement data in the CC step, which makes ViT-FC more suitable for the online application of BMS.}
\item{The pre-trained ViT-FC model obtained through the training of the battery under known working conditions can predict the SOH of the battery under unknown working conditions only by using the battery data of four charge and discharge cycles at the beginning of the battery life cycle, which greatly reduces the training cost and data acquisition cost of the SOH prediction model.}
\end{enumerate}

\section{Conclusion}
The ageing mechanism of the battery is complex, which leads to high uncertainty of SOH prediction. Therefore, good cell feature expression is the key to SOH prediction. This paper presents a ViT-FC based SOH prediction model. The feature expression of the battery can be obtained from the cycle data by ViT, and then the SOH value is regressed and predicted through the FC layer. In addition, ViT-FC can freeze the ViT model parameters, fine-tune FC layer parameters to transfer the cycle information of the battery under known working conditions, accelerate the training speed of the battery under unknown working conditions, and improve prediction accuracy. To our knowledge, this study is one of the first attempts to apply ViT to the SOH prediction of LIBs. The validity of ViT-FC is verified by the battery cycle data of 12 different working conditions. The results show that ViT-FC has good prediction accuracy and strong transfer ability.

Although the deep learning method proposed in this paper shows strong advantages, it still has some limitations. First of all, this paper only considers the acquisition of the cycle features of the battery, ignoring the temporal features of the battery ageing. We can consider using the fusion model to deepen the acquisition of features. Secondly, the data of the fixed voltage segment in the CC step is selected as the input of the whole model, without considering the influence of the length of the voltage segment on the final prediction result. The long voltage segment may affect the online application of the method. Finally, the experiments in this paper are all based on the different cycle conditions of the same type of battery, which may limit the application of this method. In the future, we will study the transfer learning methods of different types of LIBs.

\bibliographystyle{IEEEtran}
\bibliography{ref}

\begin{IEEEbiography}[{\includegraphics[width=1in,height=1.25in,clip,keepaspectratio]{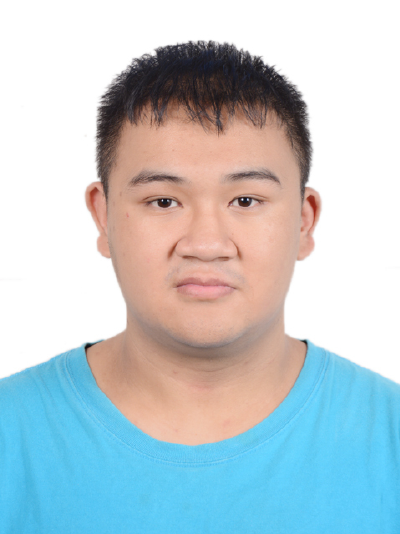}}]{Pengyu Fu} received the B.S. degree in vehicle engineering from Chongqing University, Chongqing, China, in 2019. He is currently working toward the M.S. degree in vehicle engineering from Jilin University, Changchun, China. 

His current research interests include machine learning and neural networks.
\end{IEEEbiography}

\begin{IEEEbiography}[{\includegraphics[width=1in,height=1.25in,clip,keepaspectratio]{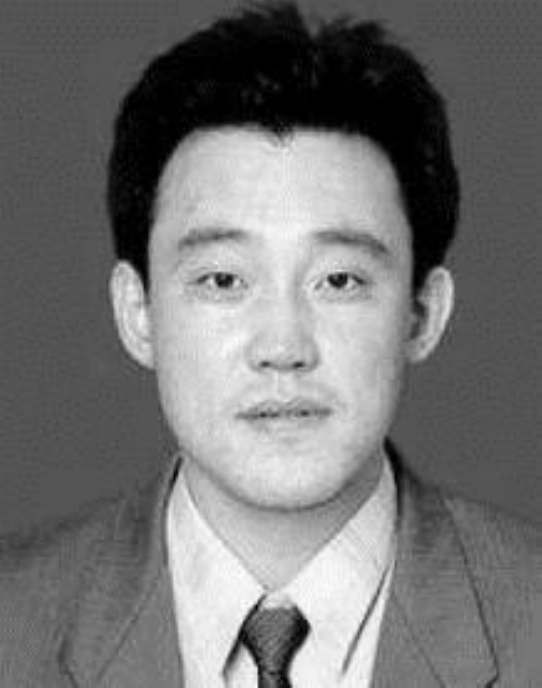}}]{Liang Chu} was born in 1967. He received the B.S., M.S., and Ph.D. degrees in vehicle engineering from Jilin University, Changchun, China. He is currently a Professor and the Doctoral Supervisor with the College of Automotive Engineering, Jilin University. His research interests include the driving and braking theory and control technology for hybrid electric vehicles, which conclude powertrain and brake energy recovery control theory and technology on electric vehicles and hybrid vehicles, theory and technology of hydraulic antilock braking and stability control for passenger cars, and the theory and technology of air brake ABS, and the stability control for commercial vehicle. 

Dr. Chu has been a SAE Member. He was a member at the Teaching Committee of Mechatronics Discipline Committee of China Machinery Industry Education Association in 2006. 
\end{IEEEbiography}

\begin{IEEEbiography}[{\includegraphics[width=1in,height=1.25in,clip,keepaspectratio]{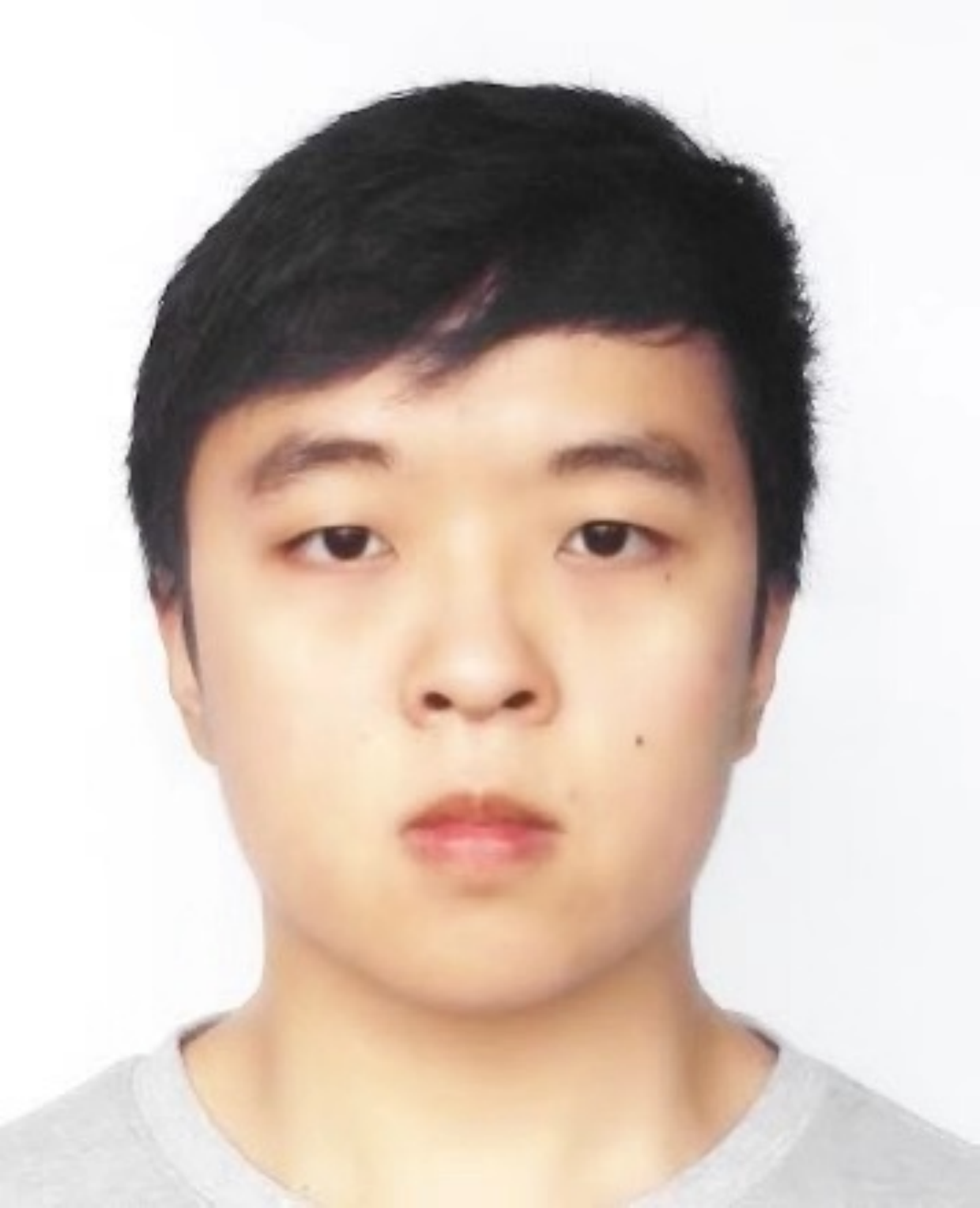}}]{Zhuoran Hou} received the B.S. degree in vehicle engineering from Chongqing University, Chongqing, China, in 2017. and the M.S. in Automotive Engineering from Jilin University, China, in 2020. He is currently pursuing continuous academic program involving doctoral studies in automotive engineering with Jilin University, Changchun, China. 

His research interests include basic machine learning, optimal energy management strategy about plug-in hybrid vehicles. 
\end{IEEEbiography}

\begin{IEEEbiography}[{\includegraphics[width=1in,height=1.25in,clip,keepaspectratio]{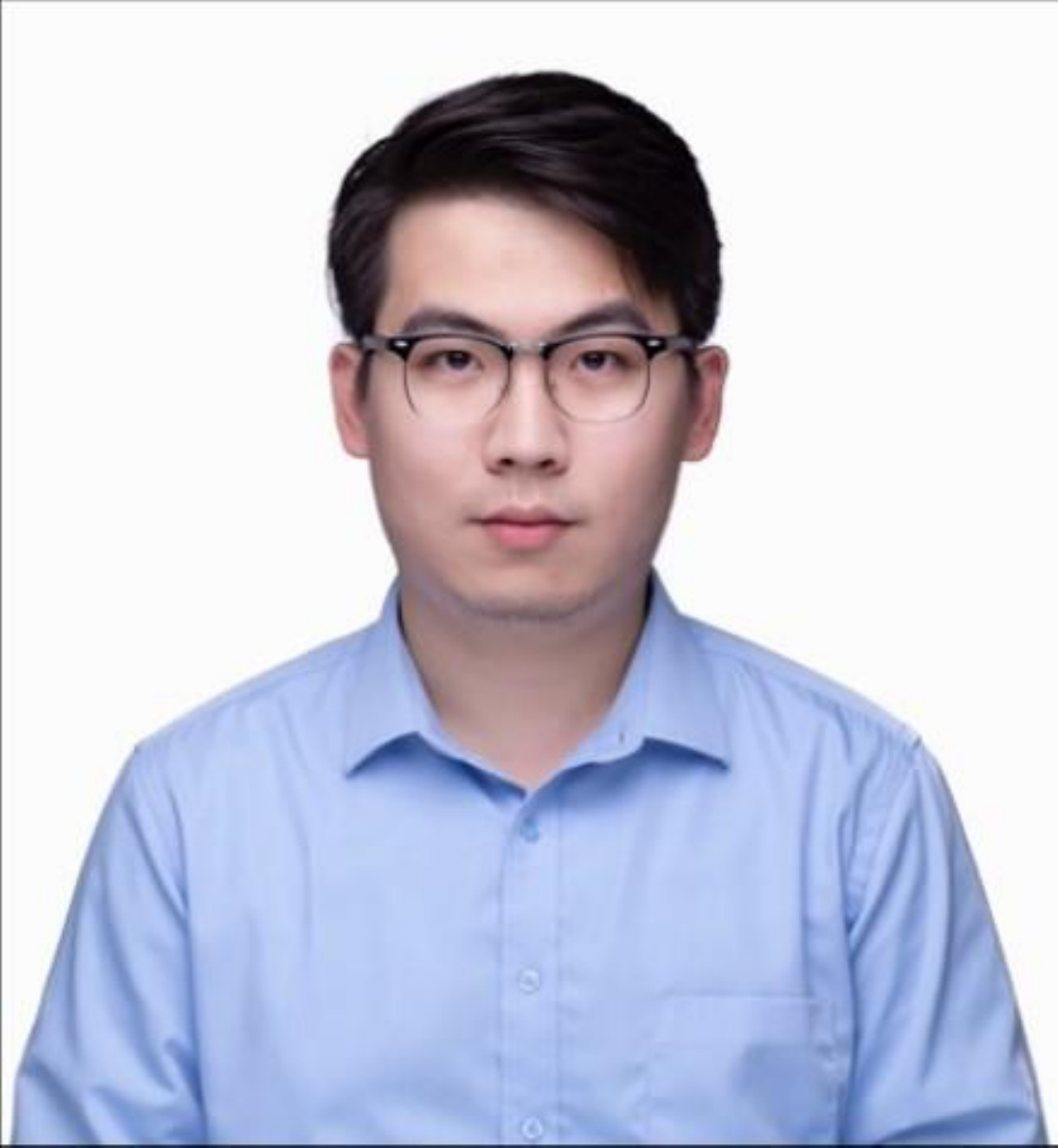}}]{Jincheng Hu} (Student Member, IEEE) received the B.E. degree in information security from the Tianjin University of Technology, Tianjin, China, in 2019 and the M. Sc. degree in information security from the University of Glasgow, Glasgow, UK, in 2022. He is currently working toward the Ph.D. degree in Automotive with the Loughborough university, Loughborough, UK. 

His research interests include reinforcement learning, deep learning, cyber security, and energy management.
\end{IEEEbiography}

\begin{IEEEbiography}[{\includegraphics[width=1in,height=1.25in,clip,keepaspectratio]{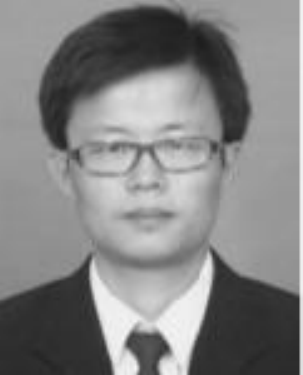}}]{Yanjun Huang} (Member, IEEE) received the Ph.D. degree in mechanical and mechatronics engineering from the University of Waterloo, Waterloo, Canada, in 2016.

He is currently a Professor with the School of Automotive Studies, Tongji University, Shanghai, China. He has authored several books and over 50 papers in journals and conferences. His research interests include the vehicle holistic control in terms of safety, energy saving, and intelligence, including vehicle dynamics and control, hybrid electric vehicle/electric vehicle optimization and control, motion planning and control of connected and autonomous vehicles, and human-machine cooperative driving. Dr. Huang serves as the Associate Editor and Editorial Board Member for the IET Intelligent Transport System, Society of Automotive Engineers (SAE) International Journal of Commercial vehicles, International Journal of Vehicle Information and Communications, Automotive Innovation, etc.
\end{IEEEbiography}

\begin{IEEEbiography}[{\includegraphics[width=1in,height=1.25in,clip,keepaspectratio]{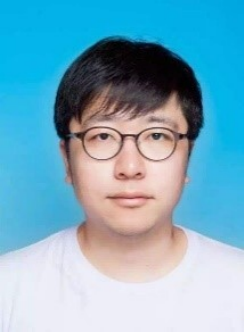}}]{Yuanjian Zhang} (Member, IEEE) received the M.S. in Automotive Engineering from the Coventry University, UK, in 2013, and the Ph.D. in Automotive Engineering from Jilin University, China, in 2018. In 2018, he joined the University of Surrey, Guildford, UK, as a Research Fellow in advanced vehicle control. From 2019 to 2021, he worked in Sir William Wright Technology Centre, Queen's University Belfast, UK. 

He is currently a Lecturer with the Department of Aeronautical and Automotive Engineering, Loughborough University, Loughborough, U.K. He has authored several books and more than 50 peer-reviewed journal papers and conference proceedings. His current research interests include advanced control on electric vehicle powertrains, vehicle-environment-driver cooperative control, vehicle dynamic control, and intelligent control for driving assist system.
\end{IEEEbiography}
 \enlargethispage{-11cm}
\end{document}